\documentclass[10pt,twocolumn,letterpaper]{article}

\usepackage{cvpr}
\usepackage{times}
\usepackage{epsfig}
\usepackage{graphicx}
\usepackage{amsmath}
\usepackage{amssymb}

\usepackage{graphics}
\usepackage{pifont}
\usepackage{color}
\usepackage{booktabs}
\usepackage{multirow}
\usepackage{subcaption}
\usepackage{gensymb}
\usepackage{bm}
\usepackage[table]{xcolor}

\usepackage[pagebackref=true,breaklinks=true,letterpaper=true,colorlinks,bookmarks=false]{hyperref}

\cvprfinalcopy 

\def\cvprPaperID{1421} 

\def\Abbrev{DRISF}
\def\ModelName{deep rigid instance scene flow}

\ifcvprfinal\fi
\begin{document}

\title{Deep Rigid Instance Scene Flow}

\author{
Wei-Chiu Ma$^{1,2}$ \quad Shenlong Wang$^{1,3}$ \quad Rui Hu$^{1}$ \quad Yuwen Xiong$^{1,3}$ \quad Raquel Urtasun$^{1,3}$\\
$^{1}$Uber Advanced Technologies Group\\$^{2}$Massachusetts Institute of Technology\\$^{3}$University of Toronto\\
}

\newcommand{\shenlong}[1]{\textcolor{blue}{S: #1}}
\newcommand{\raquel}[1]{\textcolor{red}{Raquel: #1}}
\newcommand{\weichiu}[1]{\textcolor{blue}{Wei-Chiu: #1}}
\newcommand{\todo}[1]{\textcolor{red}{To-Do: #1}}
\newcommand{\cbd}[1]{\textcolor{black}{#1}}
\newcommand{\addon}[1]{\textcolor{black}{#1}}

\newcommand{\arxiv}[1]{\vspace{0mm}}
\newcommand{\arxivscale}[1]{\scalebox{1}}

\newcommand{\bx}{\mathbf{x}}
\newcommand{\bb}{\mathbf{b}}
\newcommand{\ba}{\mathbf{a}}
\newcommand{\bo}{\mathbf{o}}
\newcommand{\bz}{\mathbf{z}}
\newcommand{\bp}{{\bm{p}}}
\newcommand{\bq}{{\bm{q}}}
\newcommand{\bn}{\mathbf{n}}
\newcommand{\bw}{\mathbf{w}}
\newcommand{\cI}{\mathcal{I}}
\newcommand{\cF}{\mathcal{F}}
\newcommand{\bK}{\mathbf{K}}
\newcommand{\bR}{\mathbf{R}}
\newcommand{\bW}{\mathbf{W}}
\newcommand{\bJ}{\mathbf{J}}
\newcommand{\cL}{\mathcal{L}}
\newcommand{\cS}{\mathcal{S}}
\newcommand{\cD}{\mathcal{D}}
\newcommand{\cR}{\mathcal{R}}
\newcommand{\cG}{\mathcal{G}}
\newcommand{\bbR}{\mathbb{R}}
\newcommand{\cM}{\mathcal{M}}
\newcommand{\by}{\mathbf{y}}
\newcommand{\ut}{^{(t)}}
\newcommand{\up}{^{(t-1)}}
\newcommand{\bt}{\mathbf{t}}
\newcommand{\bxi}{\bm{\xi}}

\maketitle
\thispagestyle{empty}

\begin{abstract}
In this paper we tackle the problem of scene flow estimation in the context of self-driving. We leverage deep learning techniques as well as strong priors as in our application domain the motion of the scene can be composed by the motion of the robot and the 3D motion of the actors in the scene. We formulate the problem as energy minimization in a deep structured model, which can be solved efficiently in the GPU by unrolling a Gaussian-Newton solver.  Our experiments in the challenging KITTI scene flow dataset show that we outperform the state-of-the-art by a very large margin, while being 800 times faster. \footnote{The uncompressed version of this paper and the supp. material can be found at \href{http://bit.ly/CVPR-DRISF}{here}}.
\end{abstract}
\section{Introduction}

Scene flow refers to the problem of estimating a three-dimenional motion field from a set of two consecutive (in time) stereo pairs. 
It was first introduced in \cite{vedula1999three} to describe the 3D motion of each point in the scene.
Through scene flow, we can gain insights into the geometry as well as the overall composition and motion of the scene.
It is of particular importance for robotics systems, such as self-driving cars, as knowing the 3D motion of other objects in the scene can not only help the autonomous systems avoid collision while planing its own future movements, but also improve the understanding of the scene and predict the intent of others. In this work, we focus on estimating the 3D scene flow in autonomous driving scenarios. 

In the world of self-driving, the motion of the scene can be mostly explained by the motion of the ego-car. The presence of dynamic objects which typically move rigidly can also be utilized as strong priors. Previous structure prediction approaches often exploit these facts and fit a piece-wise rigid representations of motion \cite{vogel2013piecewise,yamaguchi2014efficient,menze2015object,behl2017bounding}. 
While these methods achieve impressive results on scene flow estimation, they require minutes to process each frame, and thus cannot be employed in real-world robotics systems. 

\begin{figure}
\arxiv{-0.5cm}
\centering
\includegraphics[width=0.9\linewidth, trim={0 0mm 0 8mm},clip]{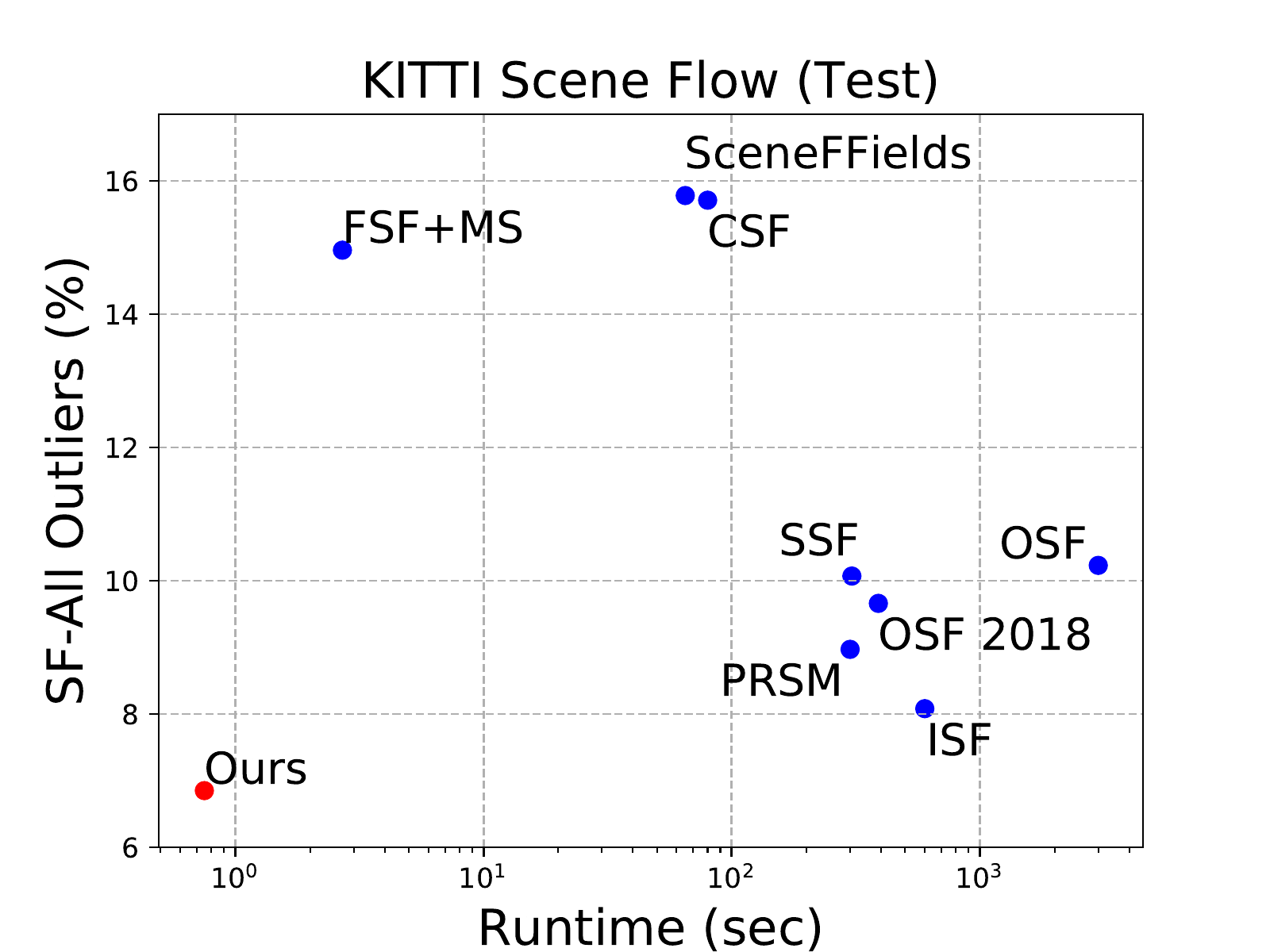}
\arxiv{-0.3cm}
\caption{\textbf{Performance vs runtime on KITTI SceneFlow dataset:} Our approach is much faster and more accurate.}
\arxiv{-0.7cm}
\label{fig:perf-vs-runtime}
\end{figure}

On the other hand, deep learning based methods have achieved state-of-the-art performance in real time on a variety of low level tasks, such as optical flow prediction \cite{fischer2015flownet,ranjan2017optical,sun2018pwc} and stereo estimation \cite{zbontar2015computing,mayer2016large,luo2016efficient}. While they produce `accurate' results, their output is not structured and cannot capture the relationships between estimated variables. For instance, they lack the ability to guarantee that pixels on a given object produce consistent estimates. While this phenomenon may have little impact in photography editing applications, this can cathastrophic in the context of self-driving cars, where the motion of the full object is more important than the motion of each individual pixel.

\begin{figure*}[!ht]
\arxiv{-0.7cm}
\centering
\includegraphics[width=1\linewidth, trim={4mm 0 7mm 0},clip]{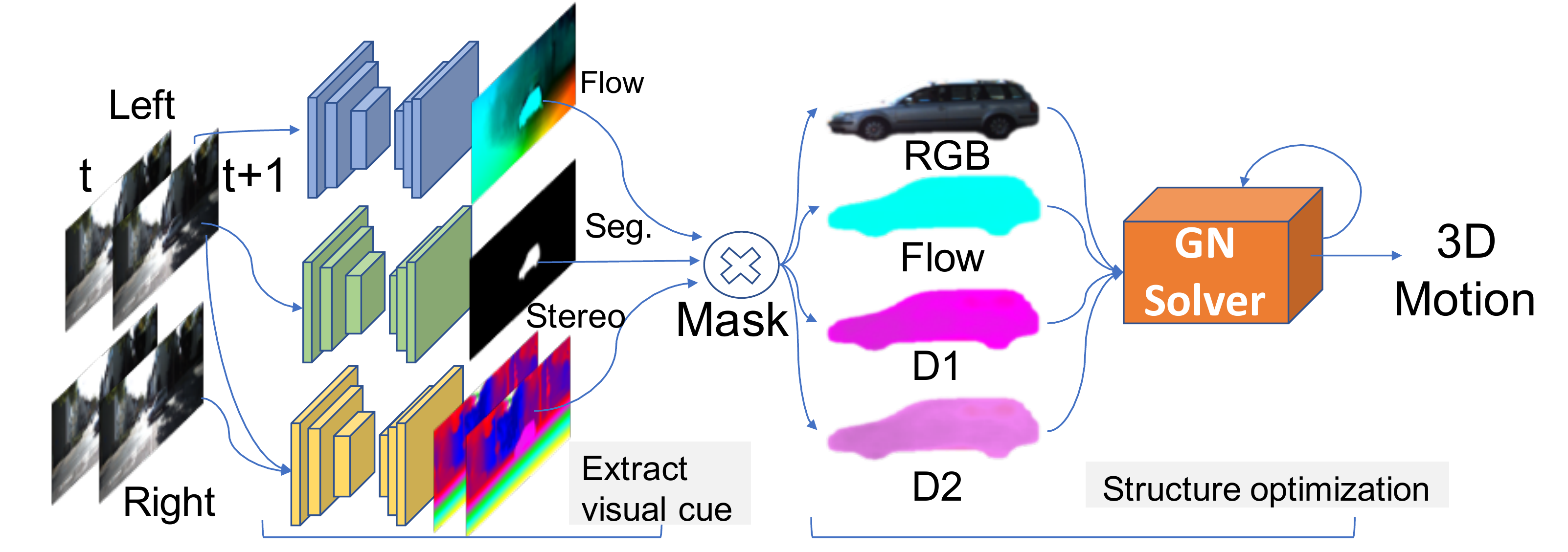}
\arxiv{-0.2cm}
\caption{\textbf{Overview of our approach:} Given two consecutive stereo images, we first estimate the flow, stereo, and segmentation (Sec. \ref{sec:visual-cues}). The visual cues of each instance are then encoded as energy functions (Sec. \ref{sec:energy}) and passed into the Gaussian-Newton (GN) solver to find the best 3D rigid motion (Sec. \ref{sec:inference}). The GN solver is unrolled as a recurrent network.}
\arxiv{-0.6cm}
\label{fig:teaser}
\end{figure*}

With these problems in mind, we develop a novel \ModelName~(\Abbrev) model that takes the best of both worlds. The idea behind is that the motion of the scene can be composed by estimating the 3D rigid motion of each actor. The static background can also be modeled as a rigidly moving object, as its 3D motion can be described by the `ego-car' motion. The problem is thus reduced to estimating the 3D motion of each traffic participant. Towards this gaol, we first capitalize on deep neural networks to  estimate optical flow, disparity and instance segmentation. We then exploit multiple geometry based energy functions to encode the structural geometric relationship between these visual cues. Through optimizing the energy function, we can effectively reason about the 3D motion of each traffic participant. As the energy takes the form of weighted sum of squares, it can be efficiently minimized via Gaussian-Newton (GN) algorithm \cite{boyd2004convex}. We implement the GN solver  as layers in neural networks, thus all operations can be computed efficiently on the GPU in an end-to-end fashion. 

We demonstrate the effectiveness of our approach on the KITTI scene flow dataset \cite{menze2015object}. As shown in Fig. \ref{fig:perf-vs-runtime}, our \ModelName~model outperforms all previous methods by a significant margin in both runtime and accuracy. Importantly, it achieves state-of-the-art performance on almost every entry. Comparing to prior art, \Abbrev~reduces the D1 outliers ratio by \textbf{43\%}, the D2 outliers ratio by \textbf{32\%}, and the flow outliers ratio by \textbf{24\%}. Comparing to the existing best scene flow model \cite{behl2017bounding}, our scene flow error is \textbf{22\%} lower and our runtime is \textbf{800} times faster.

\arxiv{-0.2cm}
\section{Related Work}
\arxiv{-0.2cm}
\paragraph{Optical flow:} Optical flow is traditionally posed as an energy minimization task. It dates back to Horn and Schunck \cite{horn1981determining} where they define the energy as a combination of a data term and a smoothness term, and adopt variational inference to solve it. Since then, a variety of improvements have been proposed \cite{brox2004high,black1996the,papenberg2006highly}.
Recently, deep learning has replaced the variational approaches. Employing deep features for matching \cite{bai2016exploiting,wang2016autoscaler} improved performance by a large margin. However, as the matching results are not dense, post-processing steps are required \cite{revaud2015epicflow}. This not only reduces the speed, but also limits the overall performance.

Pioneered by Flownet \cite{fischer2015flownet}, various end-to-end deep regression based methods have been proposed \cite{ilg2018occlusions}. Flownet2 \cite{ilg2017flownet} stacks multiple networks to iteratively refine the estimated flow and introduces a differentiable warping operation to compensate for large displacements. As the resulting network is very large, SpyNet \cite{ranjan2017optical} propose to use spatial pyramid network to handle large motions. They reduce the model size greatly, yet at the cost of degrading performance. Lite-Flownet \cite{hui2018liteflownet} and PWC-Net \cite{sun2018pwc,sun2018model} extend this idea  and incorporate the traditional pyramid processing and cost volume  concepts into the network. Comparing to previous approach, the resulting model is smaller and faster. In this work, we adapt the latest PWC-Net as our flow module.

\begin{figure*}[tb]
\centering
\setlength{\tabcolsep}{1pt}
\arxivscale{0.95}{
\begin{tabular}{ccccc}
\raisebox{7px}{\rotatebox{90}{RGB}}
\includegraphics[width=0.24\linewidth]{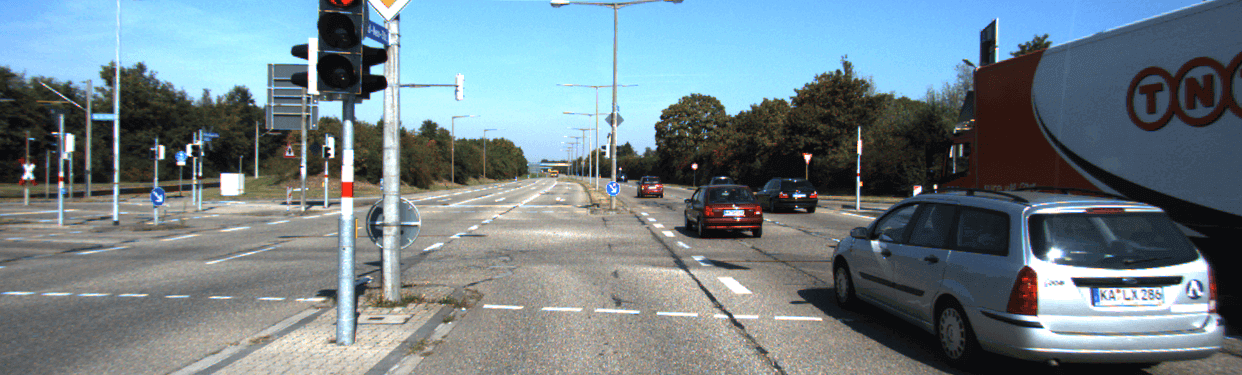}
&\includegraphics[width=0.24\linewidth]{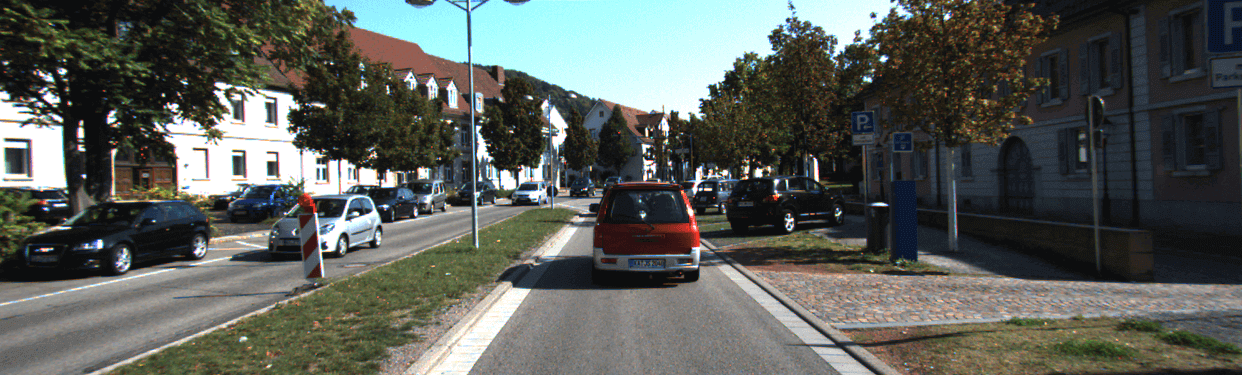}
&\includegraphics[width=0.24\linewidth]{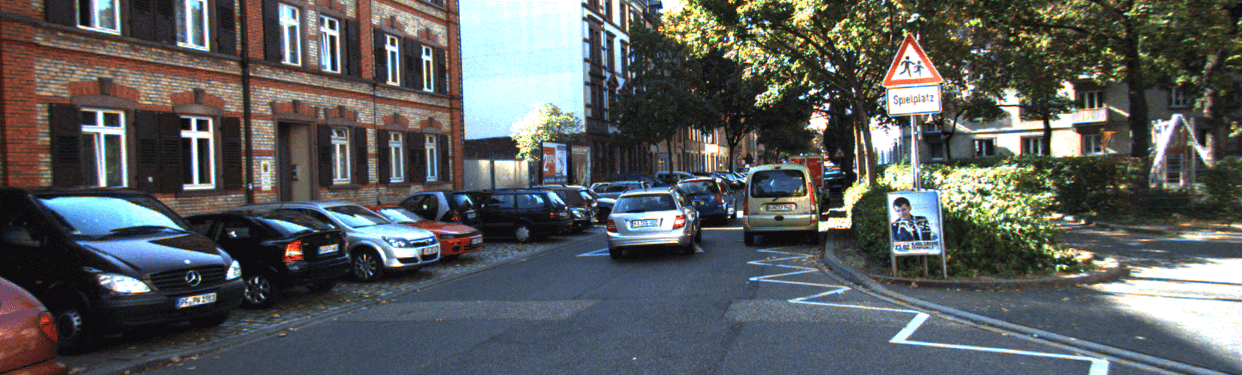}
&\includegraphics[width=0.24\linewidth]{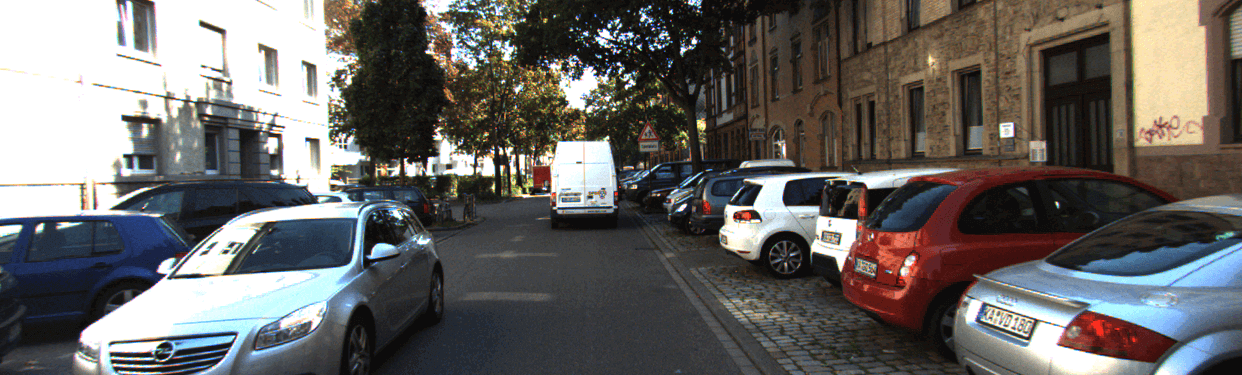}\\
\raisebox{7px}{\rotatebox{90}{RGB}}
\includegraphics[width=0.24\linewidth]{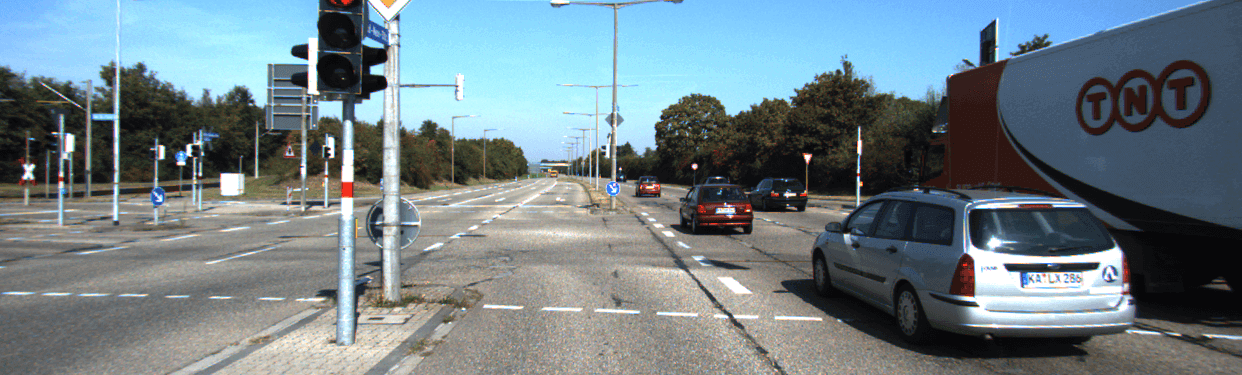}
&\includegraphics[width=0.24\linewidth]{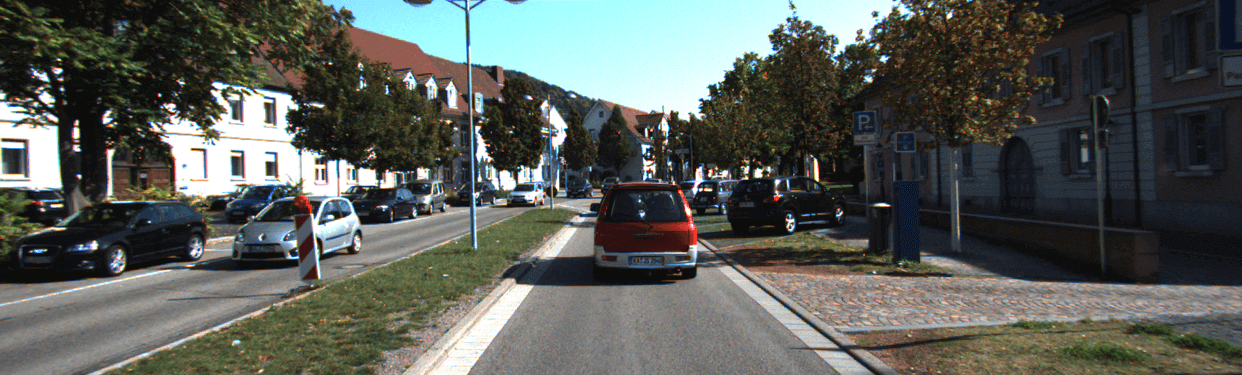}
&\includegraphics[width=0.24\linewidth]{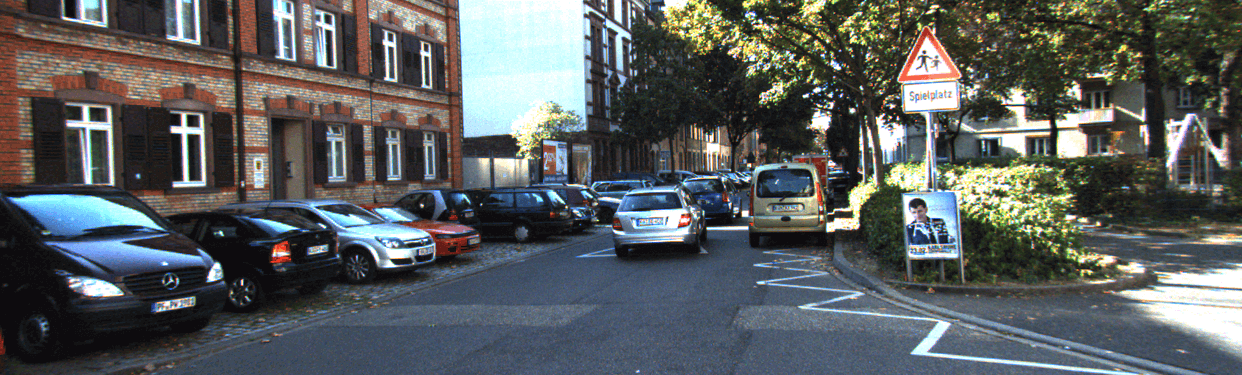}
&\includegraphics[width=0.24\linewidth]{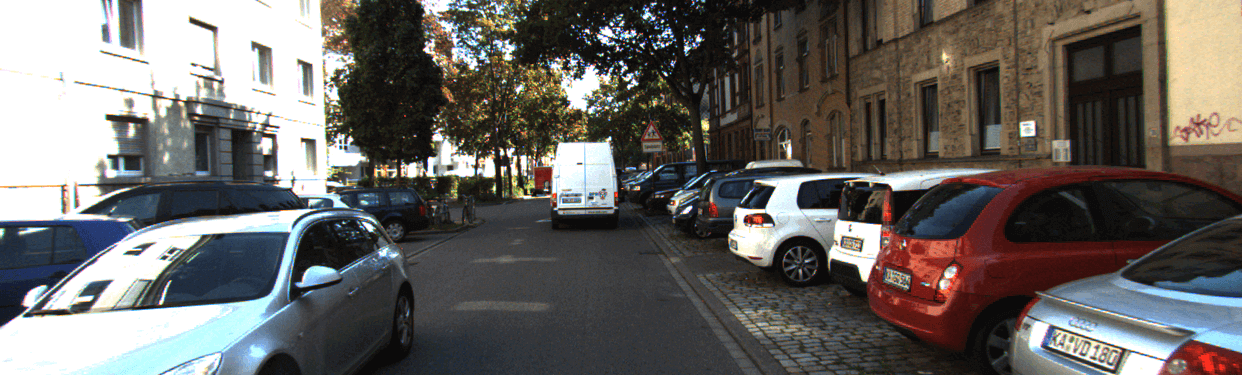}\\
\raisebox{13px}{\rotatebox{90}{D1}}
\includegraphics[width=0.24\linewidth]{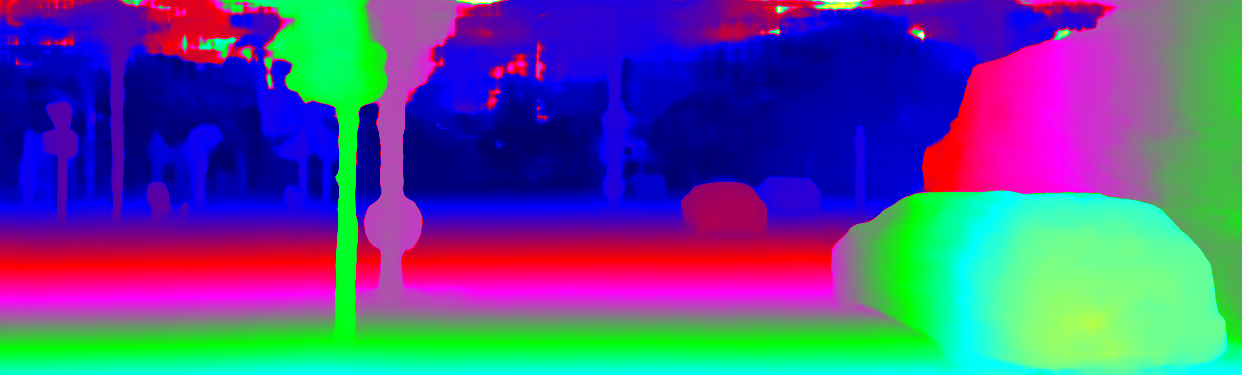}
&\includegraphics[width=0.24\linewidth]{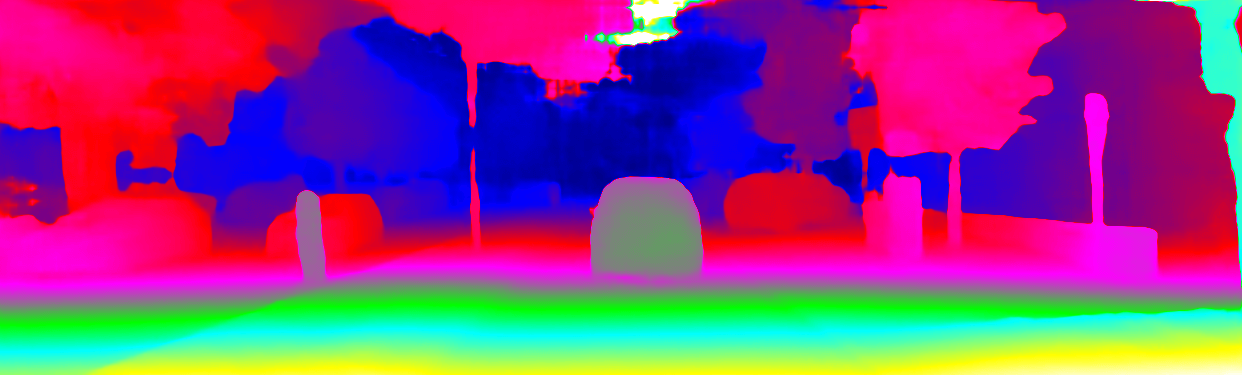}
&\includegraphics[width=0.24\linewidth]{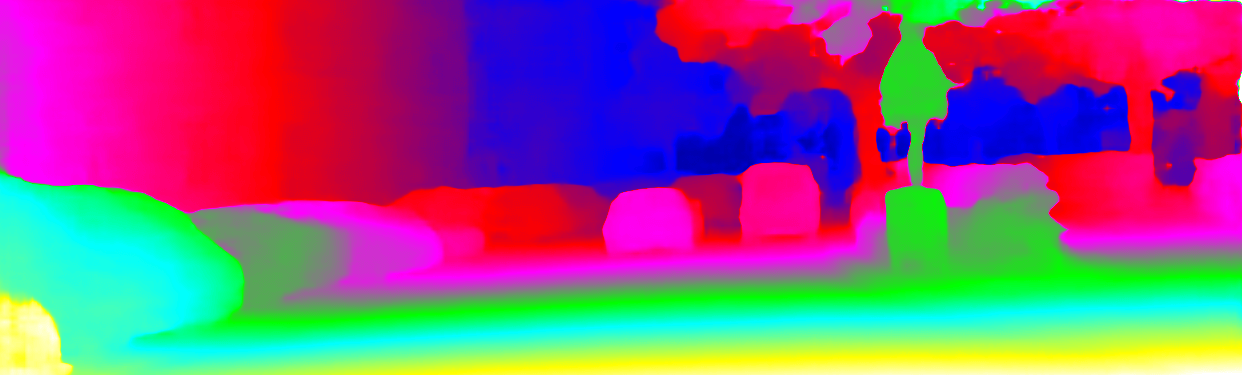}
&\includegraphics[width=0.24\linewidth]{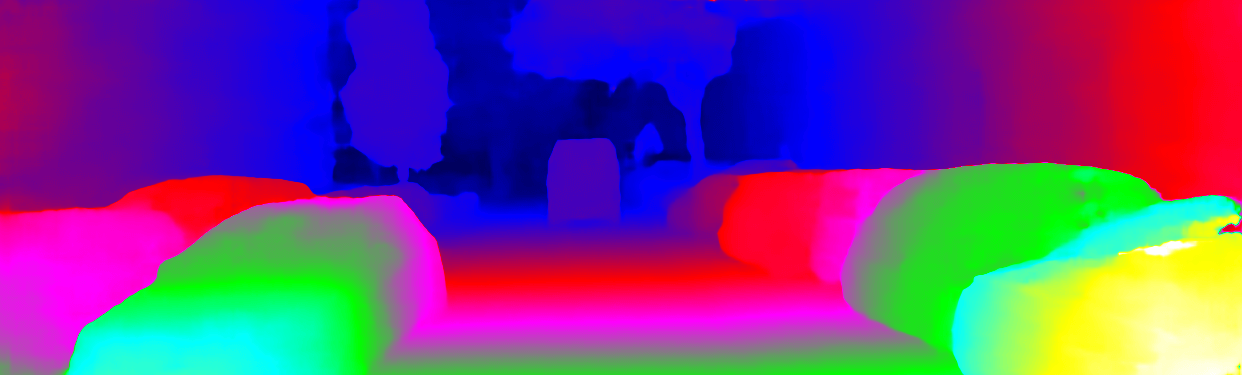}
\\
\raisebox{13px}{\rotatebox{90}{D2}}
\includegraphics[width=0.24\linewidth]{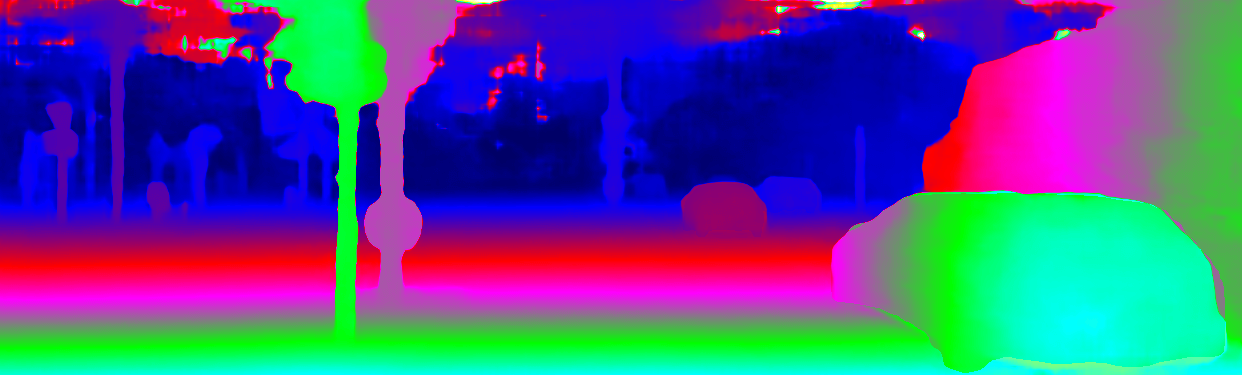}
&\includegraphics[width=0.24\linewidth]{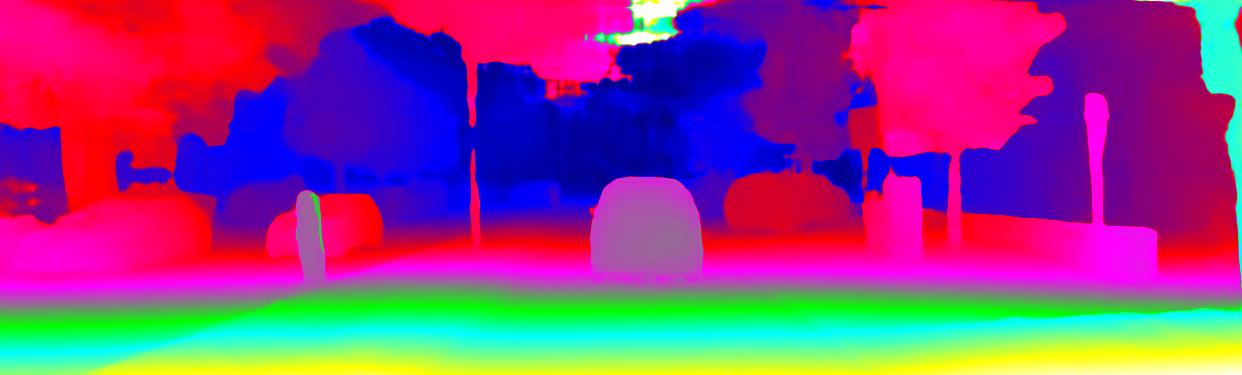}
&\includegraphics[width=0.24\linewidth]{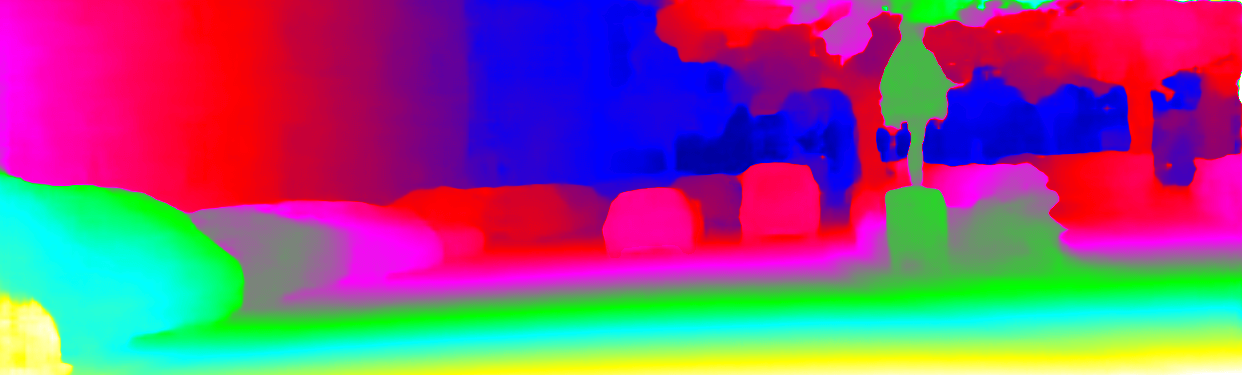}
&\includegraphics[width=0.24\linewidth]{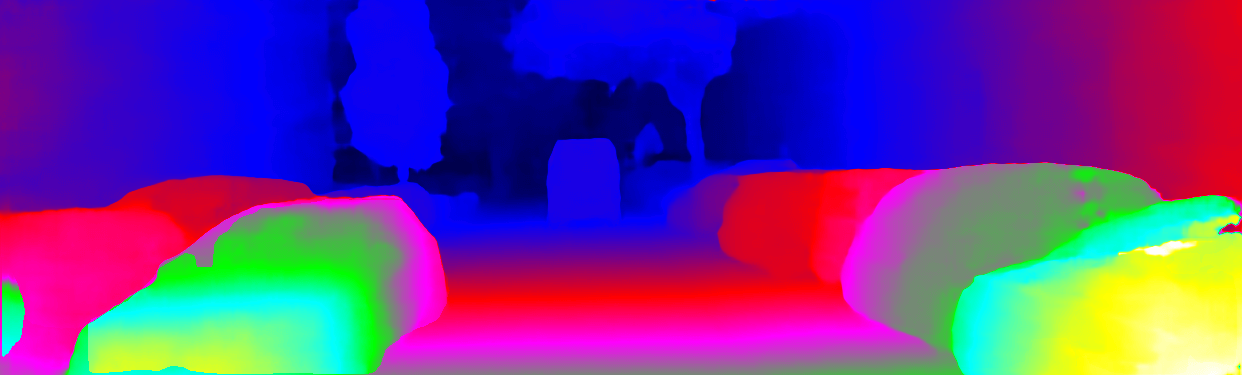}
\\
\raisebox{7px}{\rotatebox{90}{Flow}}
\includegraphics[width=0.24\linewidth]{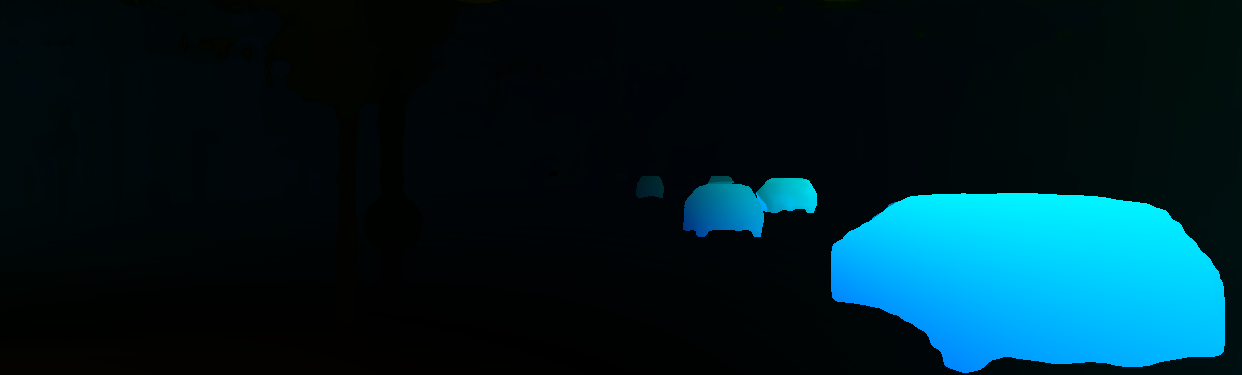}
&\includegraphics[width=0.24\linewidth]{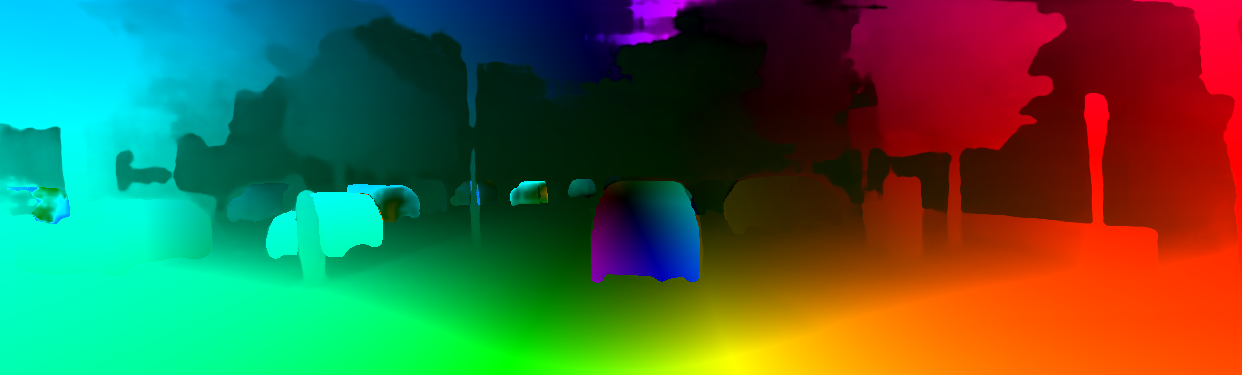}
&\includegraphics[width=0.24\linewidth]{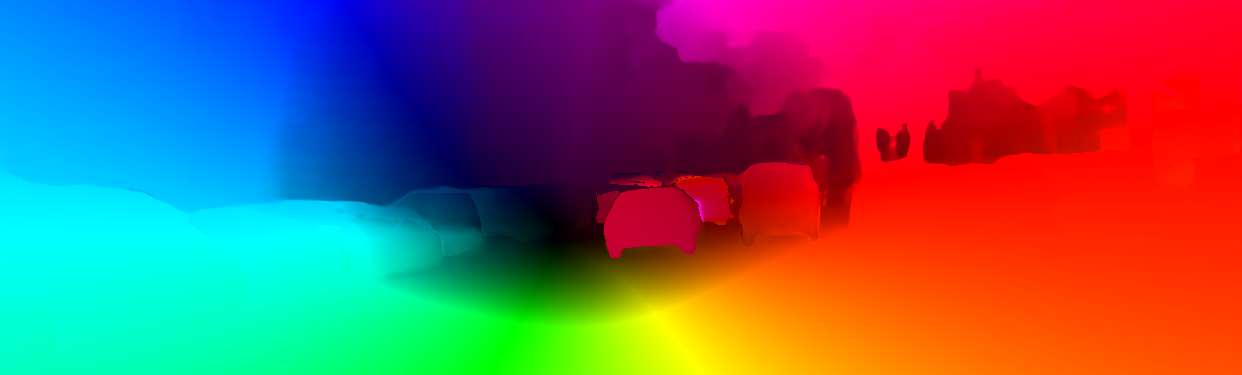}
&\includegraphics[width=0.24\linewidth]{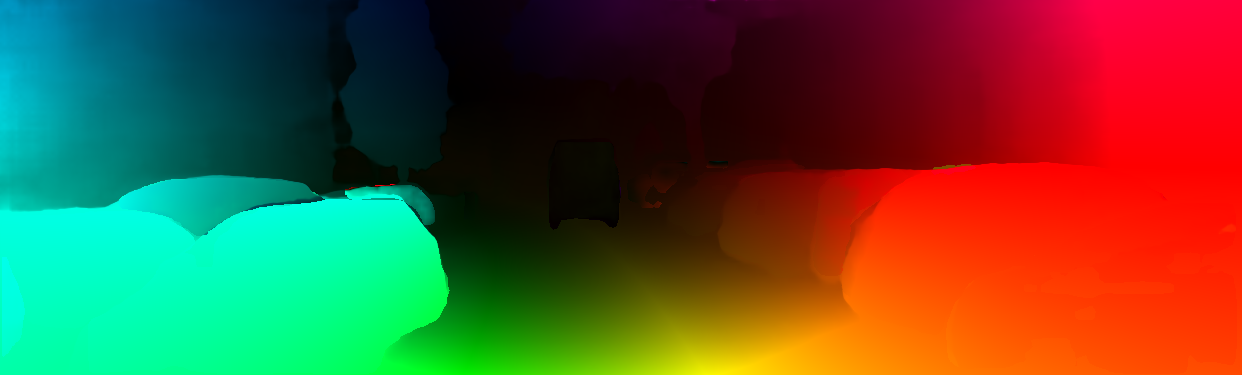}\\
\raisebox{0px}{\rotatebox{90}{SF Error}}
\includegraphics[width=0.24\linewidth]{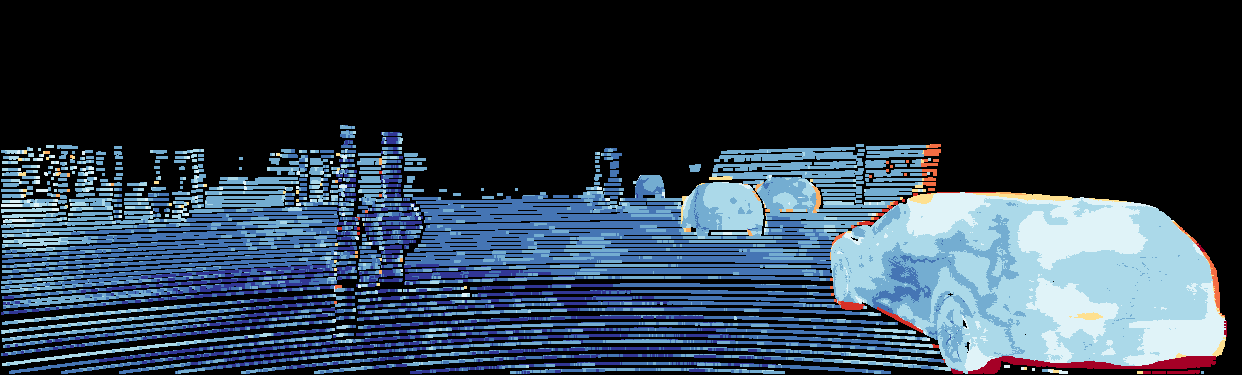}
&\includegraphics[width=0.24\linewidth]{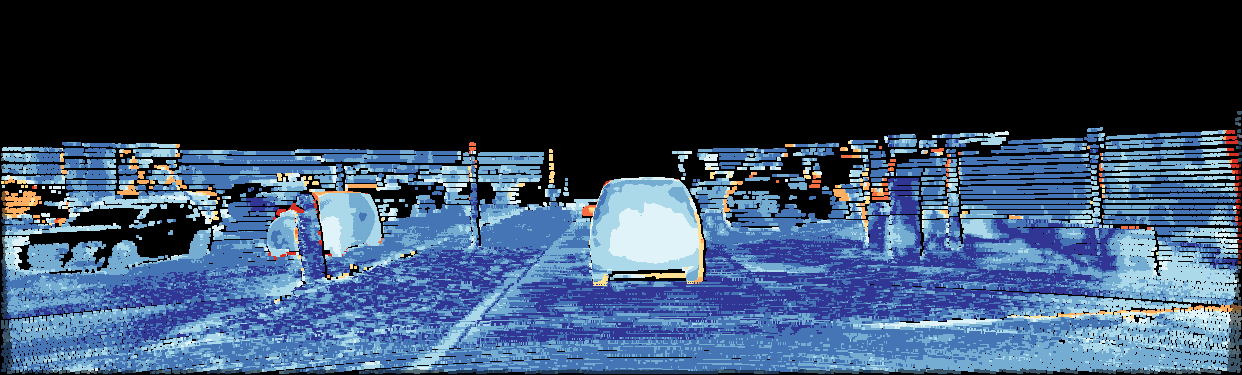}
&\includegraphics[width=0.24\linewidth]{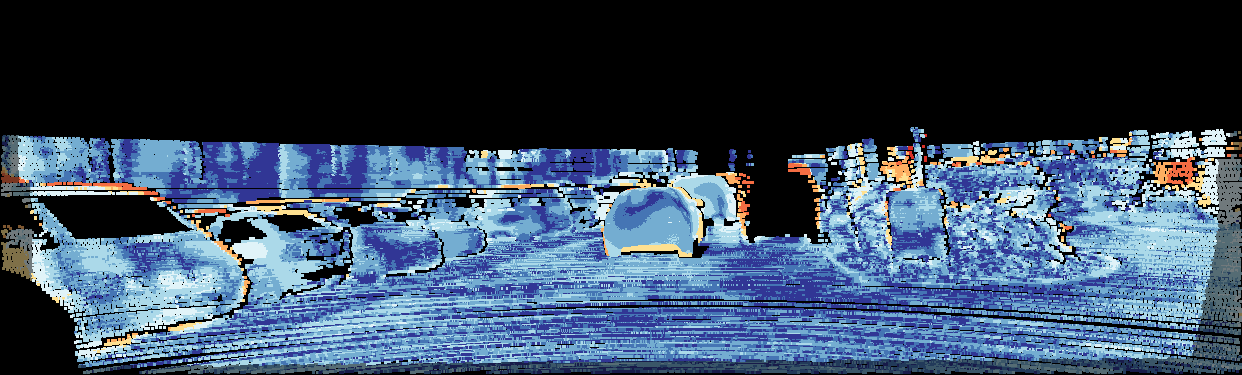}
&\includegraphics[width=0.24\linewidth]{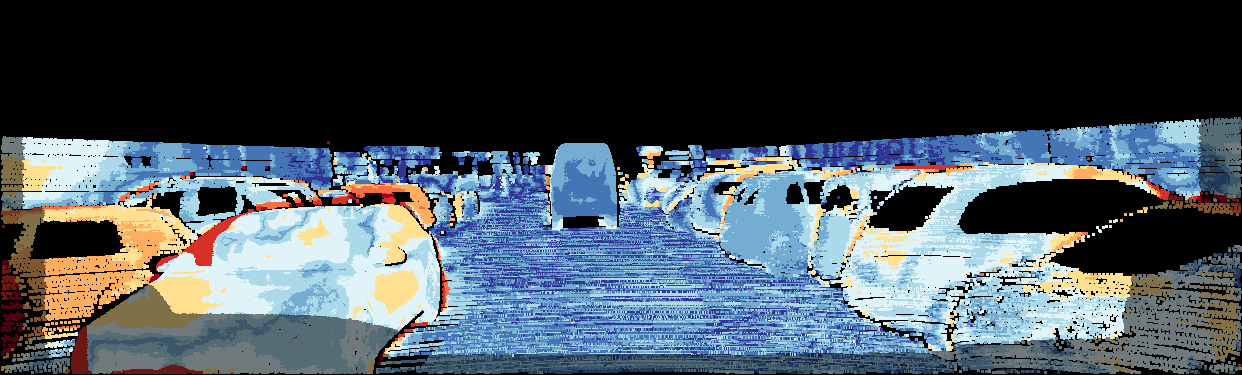}\\
\multicolumn{5}{c}{\includegraphics[width=0.99\linewidth]{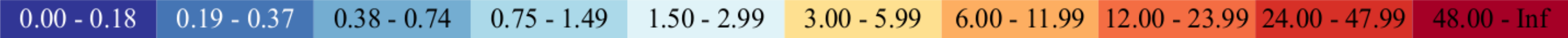}}
\\\end{tabular}
}
\arxiv{-0.4cm}
\caption{\textbf{Qualitative results on val set:} Our model can estimate the background motion very accurately. It is also able to estimate the 3D motion of foreground objects in most scenarios. It fails in challenging cases as show in last column.}
\arxiv{-0.6cm}
\label{fig:qual-results}
\end{figure*}

\arxiv{-0.5cm}
\paragraph{Stereo:}
Traditional stereo methods \cite{hoff1989surfaces,kanade1991stereo} follow three steps: compute patch-wise feature, construct cost volumes, and final post-processing. The representation of the patch  plays an important role. %
Modern approaches leverage CNNs to predict whether two patches are a match \cite{zagoruyko2015learning,zbontar2015computing}. While they showed great performance in challenging benchmarks, they are computationally expensive.
To speed up the matching process, Luo \etal \cite{luo2016efficient} propose a siamese matching network which exploits a correlation layer \cite{chen2015deep} to extract marginal distributions over all possible disparities.
While the usage of the correlation layer significantly improves efficiency, they still require post-processing techniques \cite{hirschmuller2008stereo,zbontar2016stereo} to smooth their estimation, which largely limits their speed. 
In light of this, networks that directly regress sub-pixel disparities from the given stereo image pair have been proposed. DispNet \cite{mayer2016large} exploits a 1D correlation layer to approximate the stereo cost volumes and rely on later layers for implicit aggregation. Kendall \etal \cite{kendall2017end} incorporate 3D conv for further regularization and propose a differentiable soft argmin to enable sub-pixel disparity from cost volumes. PSM-Net \cite{chang2018pyramid} later extend \cite{kendall2017end} by incorporating stacked hourglass \cite{newell2016stacked} and Pyramid spatial pooling \cite{zhao2017pyramid,he2014spatial}. In this work, we exploit PSM-Net as our stereo module.

\arxiv{-0.5cm}
\paragraph{Scene flow:} 
 Scene flow \cite{vedula1999three}  characterizes the 3D motion of a point. Similar to optical flow estimation, the task is traditionally formulated as a variational inference problem \cite{valgaerts2010joint,pons2007multi,huguet2007variational,basha2013multi}. 
However, the performance is rather limited in real world scenarios due to errors caused by large motions. To improve the robustness, slanted-plane based methods \cite{yamaguchi2014efficient,menze2015object,vogel2013piecewise,lv2016continuous} propose to decompose the scene into small rigidly moving planes and solve the discrete-continuous optimization problem.  Behl \etal \cite{behl2017bounding} build upon \cite{menze2015object}, and incorporate recognition cues. With the help of fine-grained instance and geometric feature, they are able to establish correspondences across various challenging scenarios. \addon{Similar to our work, Ren \etal \cite{ren2017cascaded} exploit multiple visual cues for scene flow estimation. They encode the features via a cascade of conditional random fields and iteratively refine them.} While these methods have achieved impressive performance, they are computationally expensive for practical usage. Most methods require minutes to compute one scene flow. This is largely due to the complicated optimization task. In contrast, our deep structured motion estimation model is able to compute scene flow in less than a second, which is two to three orders of magnitude faster.%
\arxiv{-0.3cm}
\section{Deep Rigid Instance Scene Flow}
\arxiv{-0.2cm}
In this paper we are interested in estimating scene flow in the context of self-driving cars. 
We build our model on the intuition that in this scenario the motion of the scene can be formed by estimating the 3D  motion of each actor. The static background can be also modeled as a  rigidly moving  object, as its 3D motion can be described by the `ego-car' motion.
Towards this goal, we proposed a novel  deep structured model that exploits optical flow, stereo, as well as instance segmentation as visual cues. 
We start by describing how we employ deep learning to effectively estimate the  geometric and semantic features. We then formulate the scene flow task as an energy minimization problem and discuss each energy term in details. Finally, we describe how to perform efficient inference and learning. %

\begin{table*}[tb]
\arxiv{-0.7cm}
\centering
\scalebox{0.95}{
\begin{tabular}{lccccccccccccc}
\specialrule{.2em}{.1em}{.1em}
&&\multicolumn{3}{c}{Dispairty 1}&\multicolumn{3}{c}{Dispairty 2}&\multicolumn{3}{c}{Optical Flow} &\multicolumn{3}{c}{Scene Flow}\\
Methods & Runtime &\emph{bg} &\emph{fg} &{all} &\emph{bg} &\emph{fg} &{all} &\emph{bg} &\emph{fg} &{all} &\emph{bg} &\emph{fg} &{all}\\
\hline
CSF \cite{lv2016continuous} &1.3 mins&4.57&13.04&5.98&7.92&20.76&10.06&10.40&25.78&12.96&12.21&33.21&15.71\\
OSF \cite{menze2015object} &50 mins&4.54&12.03&5.79&5.45&19.41&7.77&5.62&18.92&7.83&7.01&26.34&10.23\\
SSF \cite{ren2017cascaded} &5 mins& 3.55& 8.75& 4.42& 4.94& 17.48& 7.02& 5.63& 14.71& 7.14& 7.18& 24.58& 10.07\\
OSF-TC* \cite{artnerobject} &50 mins &4.11&9.64&5.03&5.18&15.12&6.84&5.76&13.31&7.02&7.08&20.03&9.23\\
PRSM* \cite{vogel20153d} & 5 mins &3.02&10.52&4.27&5.13&15.11&6.79&5.33&13.40&6.68&6.61&20.79&8.97\\
ISF \cite{behl2017bounding} & 10 mins&4.12&6.17&4.46&4.88&11.34&5.95&5.40&{\bf 10.29}&6.22&6.58&{\bf15.63}&8.08\\
Our \Abbrev &{\bf0.75 sec} &{\bf2.16}&{\bf4.49}&{\bf2.55}&{\bf2.90}&{\bf9.73}&{\bf4.04}&{\bf3.59}&{10.40}&{\bf4.73}&{\bf4.39}&15.94&{\bf6.31}\\
\specialrule{.1em}{.05em}{.05em}
\end{tabular}
}
\arxiv{-0.2cm}
\caption{\textbf{Comparison against top 6 published approaches:} Our method acheives state-of-the-art performance on almost every entry while being two to three orders of magnitude faster. \addon{(*: Method uses more than two temporally adjacent images.)}}
\arxiv{-0.7cm}
\label{tab:quant}
\end{table*}

\arxiv{-0.2cm}
\subsection{Visual Cues}
\label{sec:visual-cues}
\arxiv{-0.2cm}
We exploit three types of visual cues: instance segmentation, optical flow and stereo. %
\arxiv{-0.4cm}
\paragraph{Instance Segmentation:} 
We utilize Mask R-CNN \cite{he2017mask} as our instance segmentation network, as it produces state-of-the-art results in autonomous driving benchmarks such as KITTI \cite{Geiger2012CVPR} and Cityscapes \cite{Cordts2016Cityscapes}. 
Mask R-CNN is a proposal based two stage network built upon Faster R-CNN \cite{ren2015faster}.  For each object proposal, it predicts the object class, regresses its 2D box, and infers the bg/fg segmentation mask. 
\arxiv{-0.5cm}
\paragraph{Stereo:} 
We exploit the pyramid stereo matching network (PSM-Net) \cite{chang2018pyramid} to compute our stereo estimates.  
It consists of three main modules: fully convolutional feature module, spatial pyramid pooling \cite{he2014spatial,zhao2017pyramid} and 3D cost volume processing. The feature module computes a high-dimensional feature map in a fully convolutional manner; the spatial pyramid pooling aggregates context in different scales and locations to construct the cost volume; the 3D cost volume module then performs implicit cost volume aggregation and regularizes it using stacked hourglass networks. Compared to previous disparity regression networks, PSM-Net learns to refine and produce sharp disparity images that respect object boundaries better. This is of crucial importance as over-smoothed results  often deteriorates  motion estimation.
\arxiv{-0.7cm}
\paragraph{Optical Flow:} 
Our flow module is akin to PWC-Net \cite{sun2018pwc}, which is a state-of-the-art  flow network  designed based on  three classical principles (similar to stereo networks):
pyramidal feature processing, warping, and cost volume reasoning. Pyramidal feature processing encode   visual features with large context; the progressive warping reduces the cost of building cost-volume through a coarse-to-fine scheme. Cost volume reasoning further boost performance by sharpening the boundaries. We implement PWC-net with one modification: during the warping operation, we use  the feature of the nearest boundary pixel to pad if the sampling point falls outside the image, rather than 0. Empirically we found this to improve  performance. 

\subsection{Energy Formulation}
\label{sec:energy}

We now describe the energy formulation of our deep structured model. 
Let $\cL^0, \cR^0,  \cL^1, \cR^1$ be the input stereo pairs captured from two consecutive time steps. 
Let $\cD^0, \cD^1$ be the estimated stereo, and $\cF_\cL, \cF_\cR$ be the inferred flow.
Denote $\cS_\cL^0 $ as the instance segmentation computed on the left image $\cL^0$.
Assume all cameras are pre-calibrated with known intrinsics.  
We parametrize the 3D rigid motion with $\bm{\xi} \in \mathfrak{se}(3)$, the Lie-algebra associated with $SE(3)$. We use this parametrization as it is a minimal representation for 3D motion. 
For each instance $i \in \cS_\cL^0 $, we aim to find the rigid 3D motion that minimizes the weighted combination of photometric error, rigid fitting and flow consistency, where the weights are denoted as $\lambda_{\cdot, i}$. For simplicity, let $\cI = \{\cL^0, \cR^0,  \cL^1, \cR^1, \cD^0, \cD^1, \cF_\cL, \cF_\cR\}$ be input images and visual cues. We denote the set of pixels belonging to instance $i$ as $P_i = \{ \bp | S_\cL^0(\bp) = i \}$. Note that background can be considered as an `instance' since all the pixels in it undergo the same rigid transform. We obtain the 3D motion of each instance  by minimizing
\begin{align}
\min_{\bm{\xi}} 
\{ \lambda_{\text{photo},i} E_{\text{photo}, i}(\bm{\xi}; \cI) &+ \lambda_{\text{rigid},i} E_{\text{rigid}, i}(\bm{\xi}; \cI) \\
&+ \lambda_{\text{flow},i} E_{\text{flow}, i}(\bm{\xi}; \cI)\} \nonumber
\end{align}
The three energy terms are complementary. They capture the  geometry and appearance agreement between the observations and inferred rigid motion.  
Next, we describe the energy terms in more details. 

\begin{figure*}[tb]
\arxiv{-0.7cm}
\centering
\setlength{\tabcolsep}{1pt}
\arxivscale{0.95}{
\begin{tabular}{ccccc}
\raisebox{7px}{\rotatebox{90}{RGB}}
\includegraphics[width=0.24\linewidth]{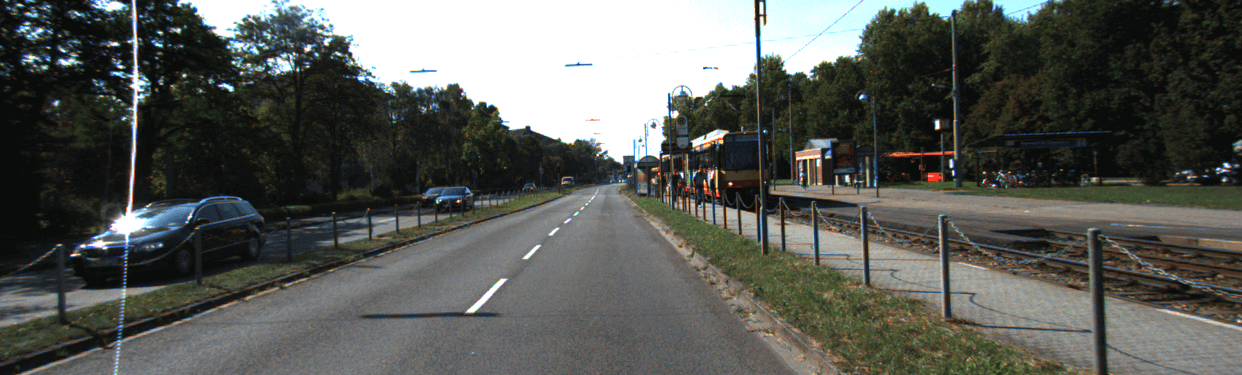}
&\includegraphics[width=0.24\linewidth]{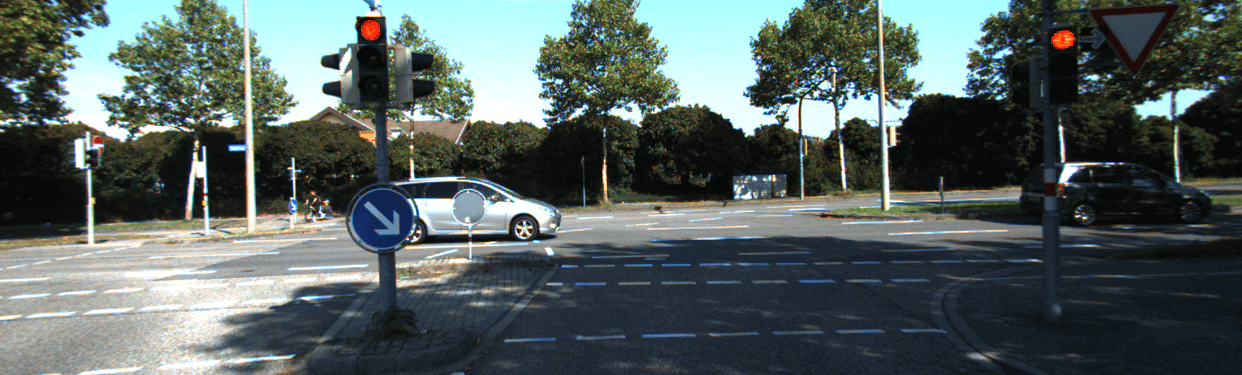}
&\includegraphics[width=0.24\linewidth]{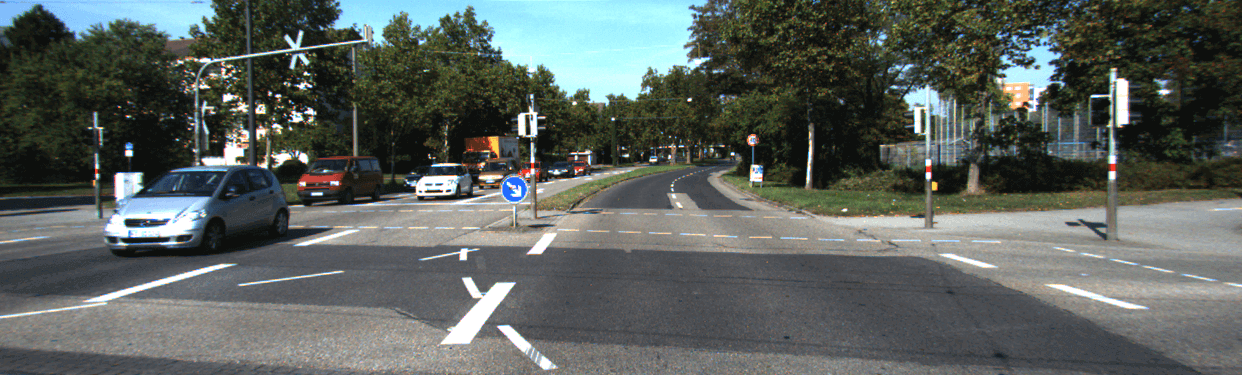}
&\includegraphics[width=0.24\linewidth]{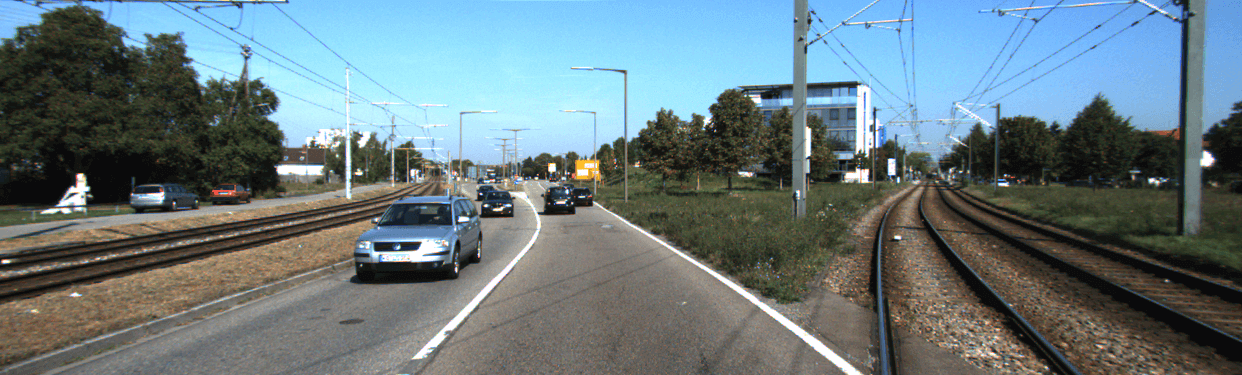}\\
\raisebox{7px}{\rotatebox{90}{RGB}}
\includegraphics[width=0.24\linewidth]{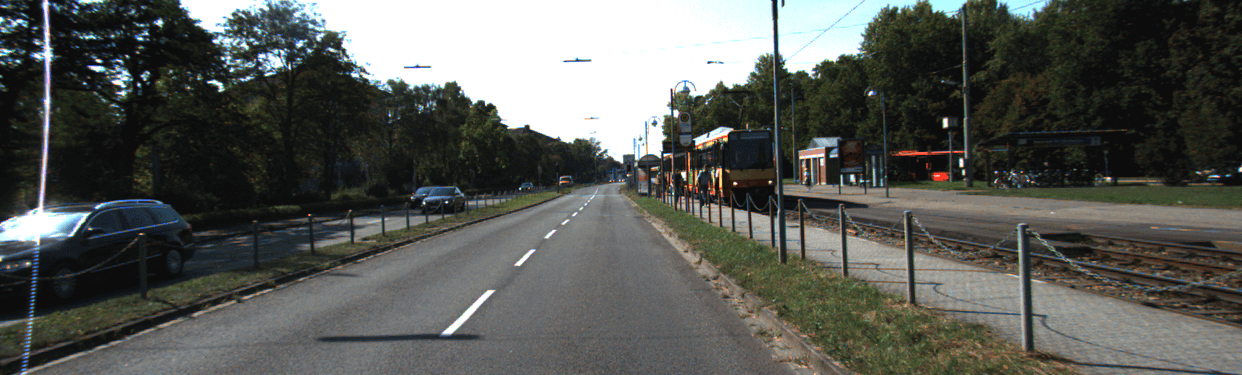}
&\includegraphics[width=0.24\linewidth]{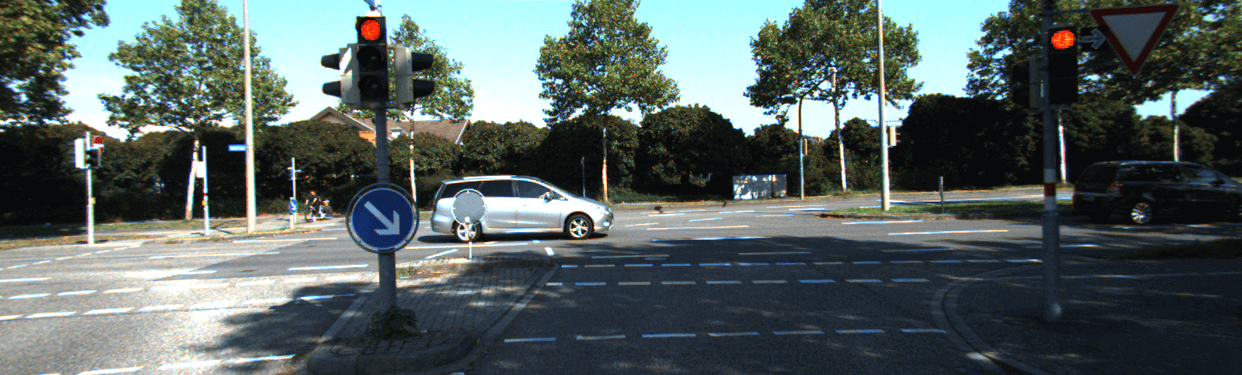}
&\includegraphics[width=0.24\linewidth]{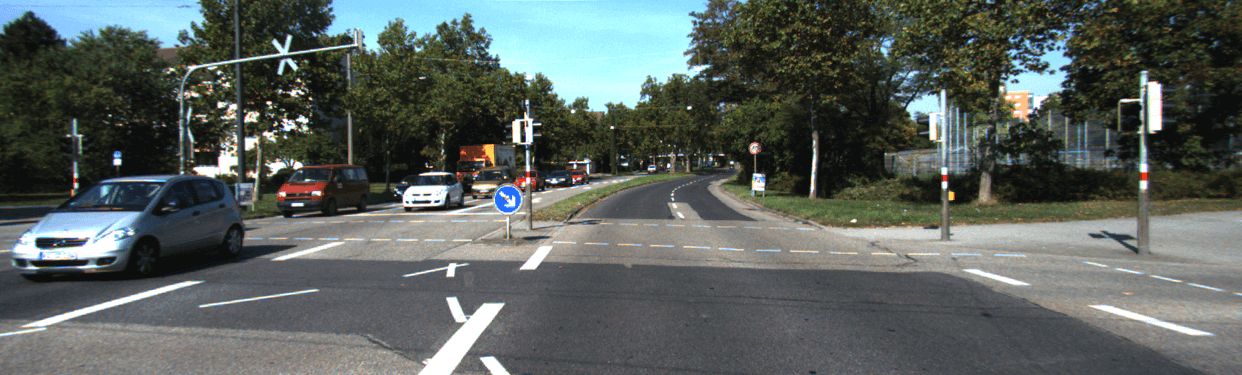}
&\includegraphics[width=0.24\linewidth]{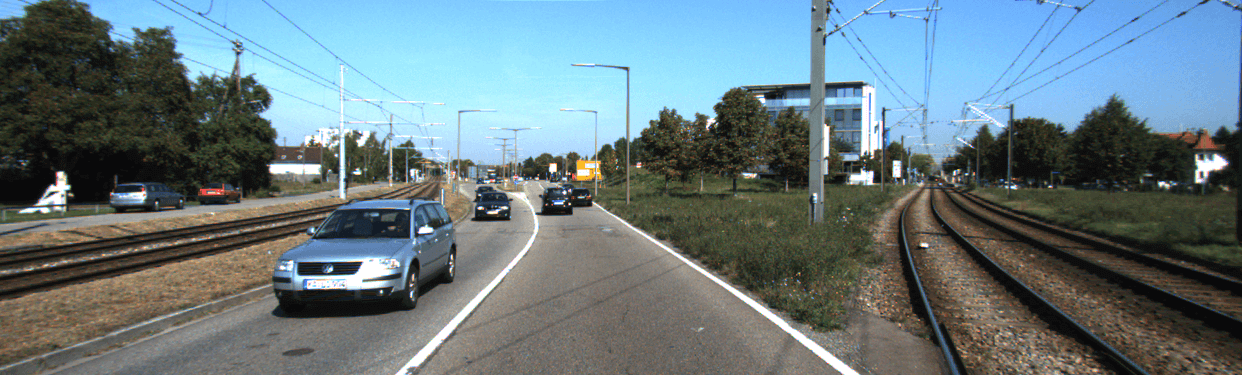}\\
\raisebox{7px}{\rotatebox{90}{OSF}}
\includegraphics[width=0.24\linewidth]{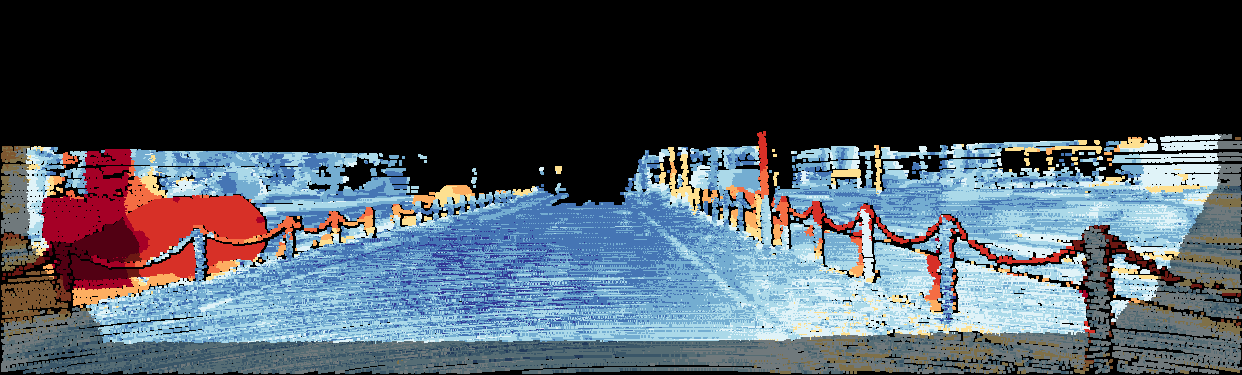}
&\includegraphics[width=0.24\linewidth]{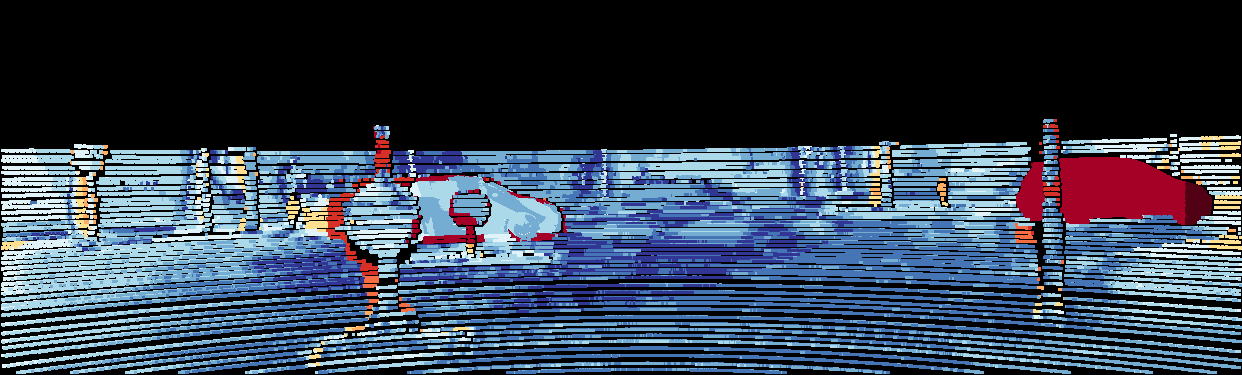}
&\includegraphics[width=0.24\linewidth]{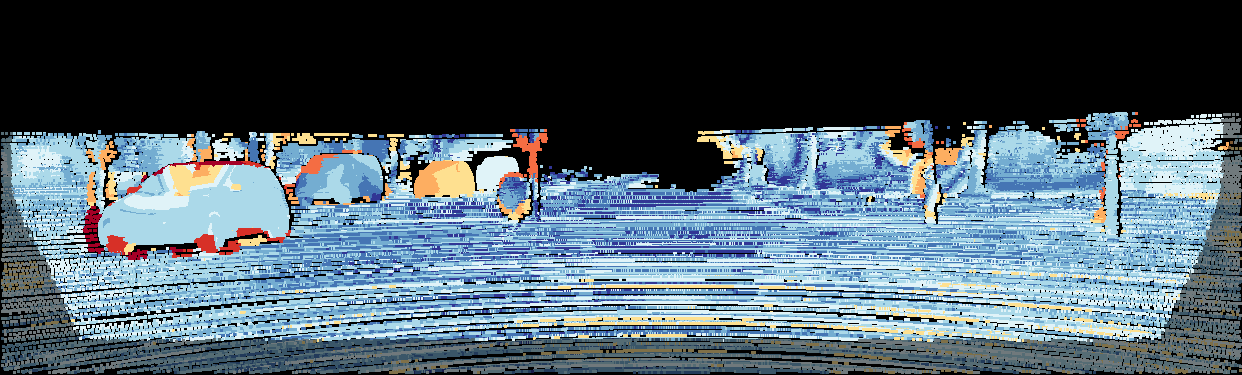}
&\includegraphics[width=0.24\linewidth]{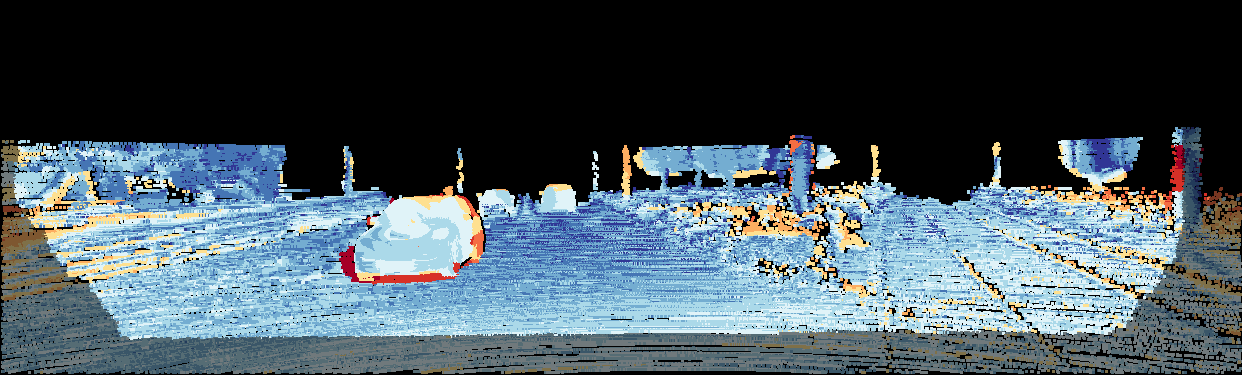}\\
\raisebox{7px}{\rotatebox{90}{PRSM}}
\includegraphics[width=0.24\linewidth]{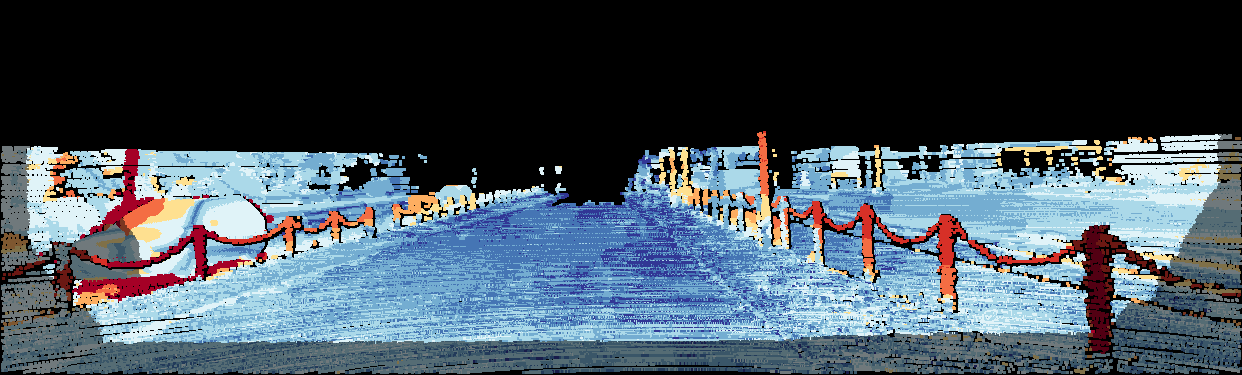}
&\includegraphics[width=0.24\linewidth]{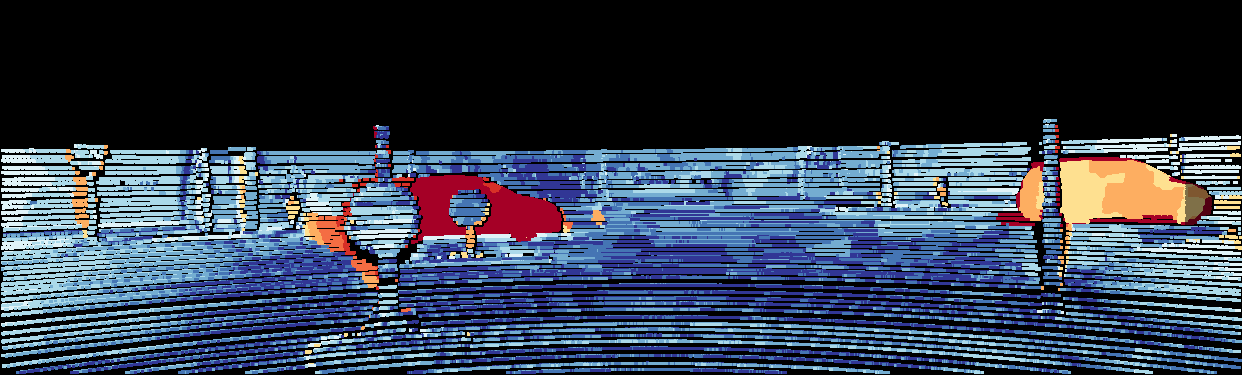}
&\includegraphics[width=0.24\linewidth]{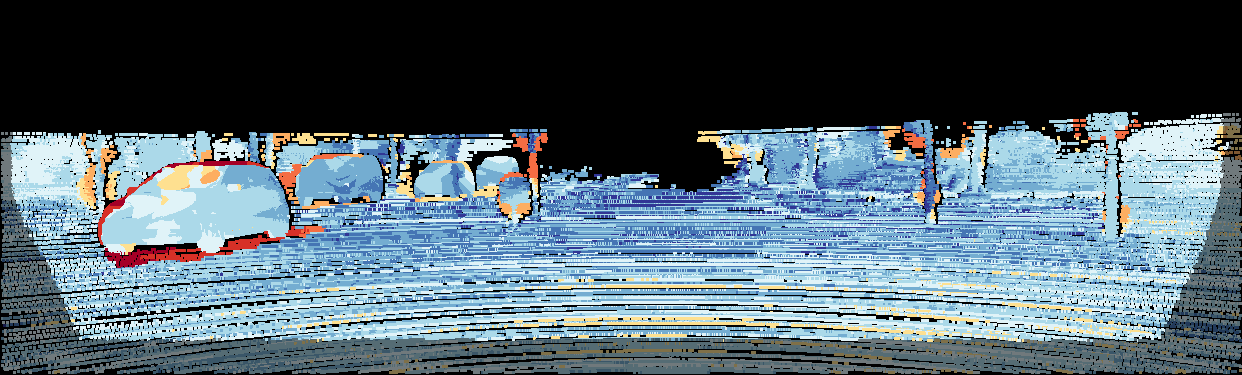}
&\includegraphics[width=0.24\linewidth]{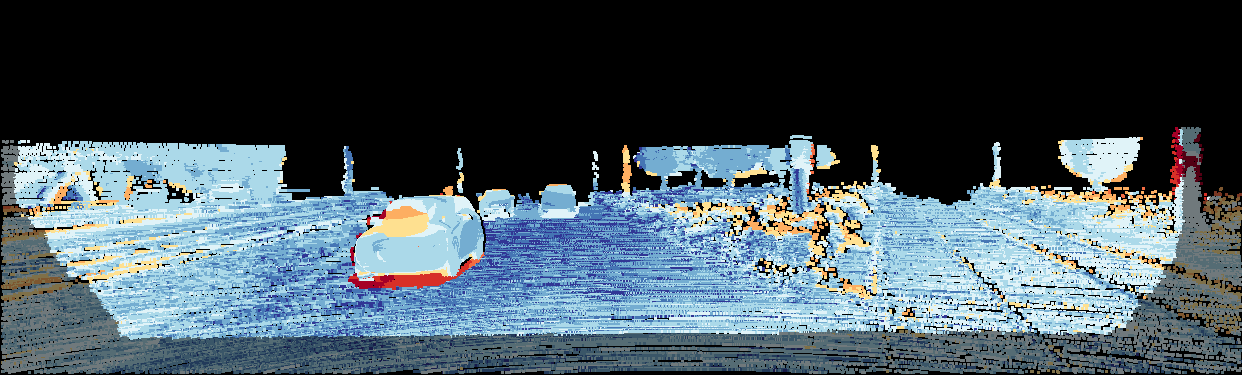}\\
\raisebox{14px}{\rotatebox{90}{ISF}}
\includegraphics[width=0.24\linewidth]{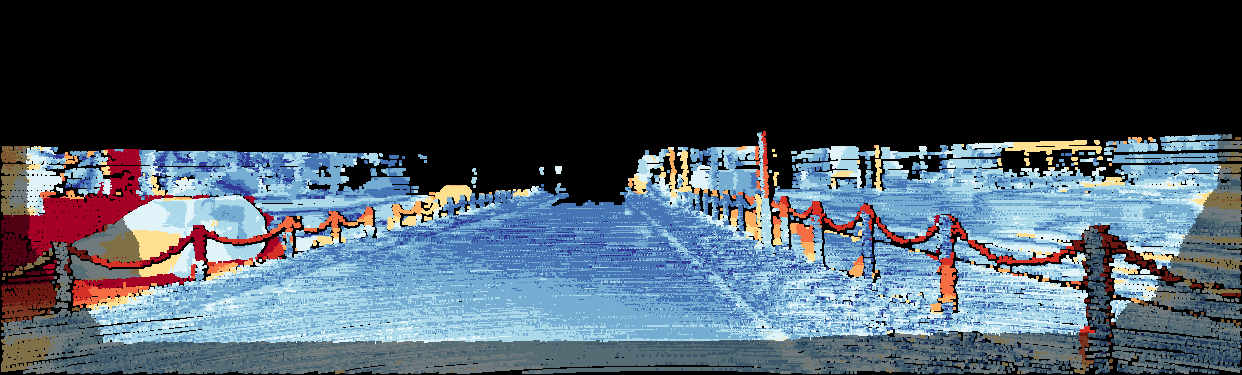}
&\includegraphics[width=0.24\linewidth]{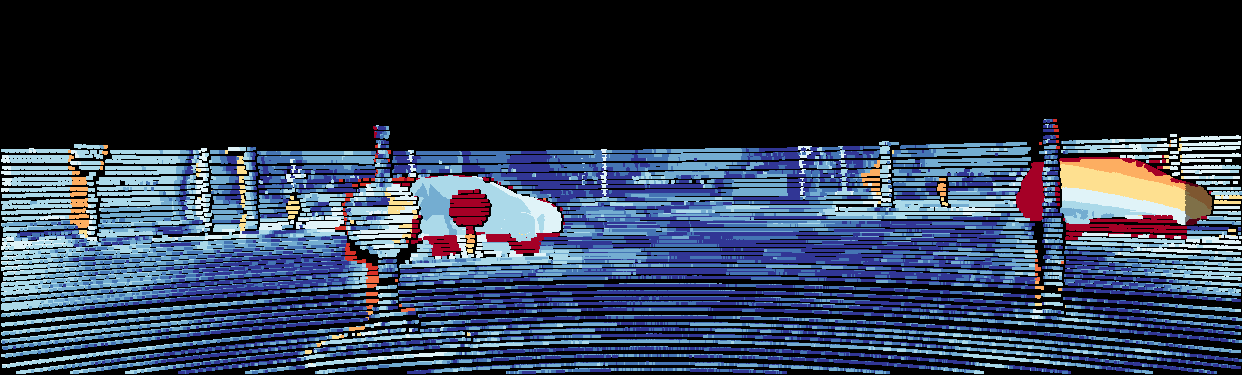}
&\includegraphics[width=0.24\linewidth]{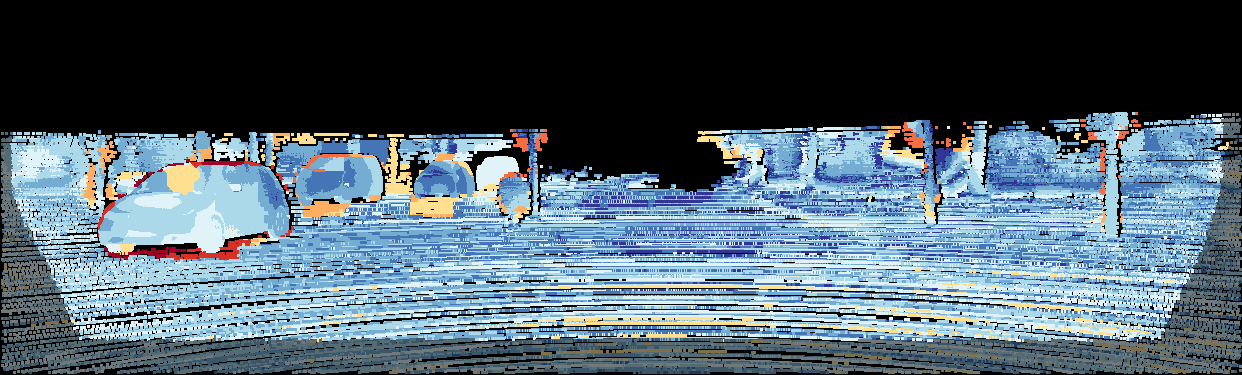}
&\includegraphics[width=0.24\linewidth]{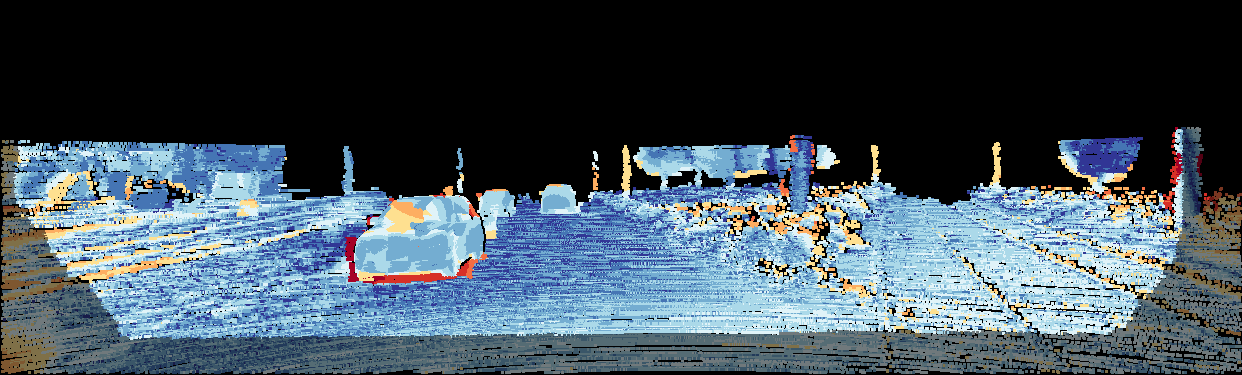}\\
\raisebox{5px}{\rotatebox{90}{\Abbrev}}
\includegraphics[width=0.24\linewidth]{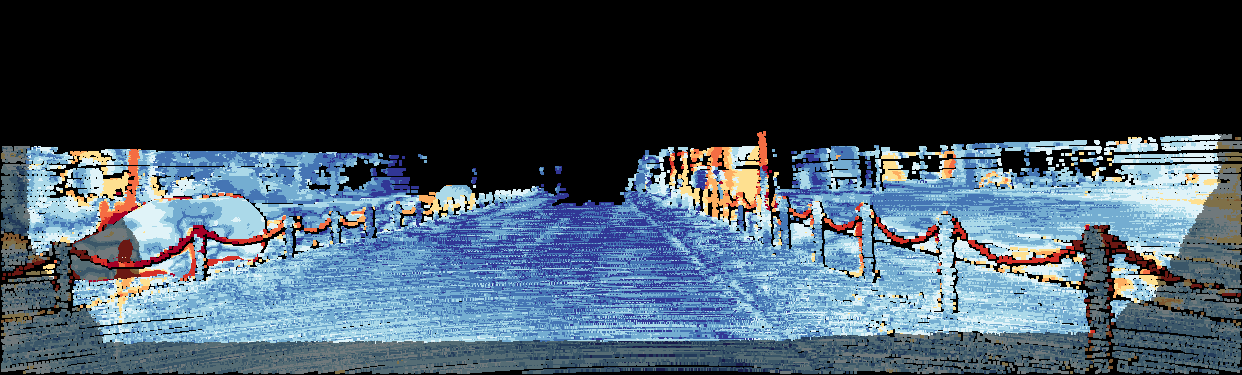}
&\includegraphics[width=0.24\linewidth]{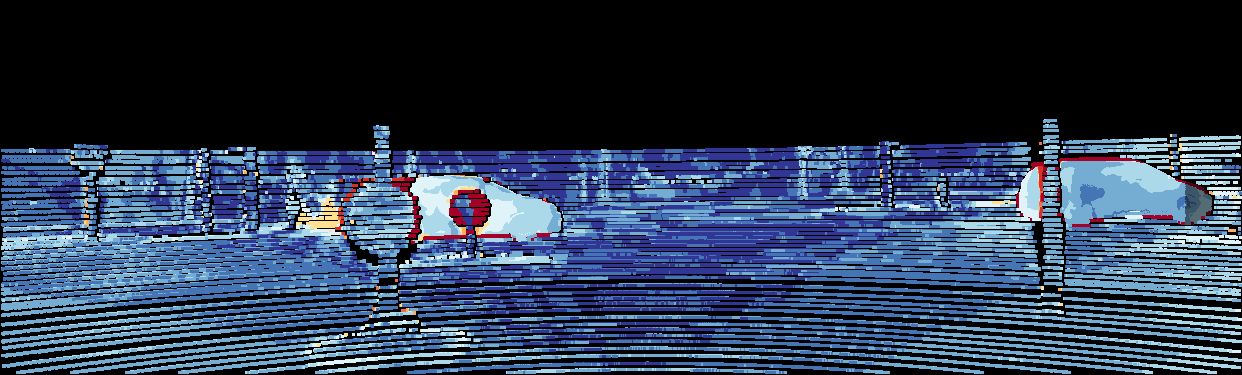}
&\includegraphics[width=0.24\linewidth]{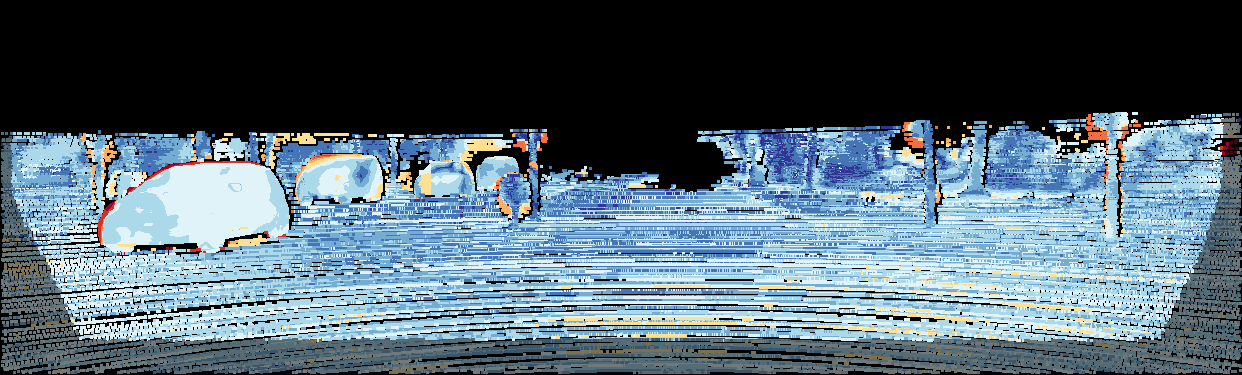}
&\includegraphics[width=0.24\linewidth]{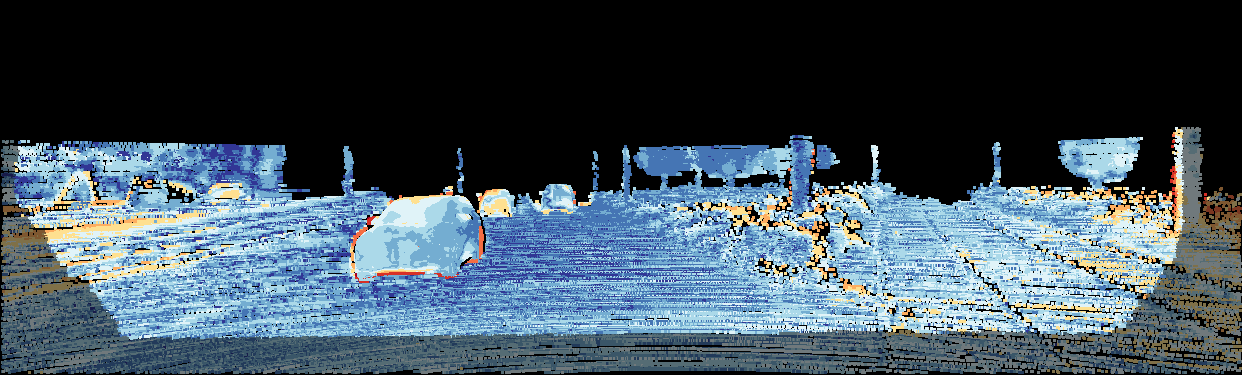}\\
\multicolumn{5}{c}{\includegraphics[width=0.99\linewidth]{img/kitti_test/error_bar.png}}
\\\end{tabular}
}
\arxiv{-0.2cm}
\caption{\textbf{Qualitative comparison on test sest:}  Our method can effecitvely handle occlusion and texture-less regions. It is more robust to the illumination change as well as large displacement. Please refer to the supp. material for more results.}
\arxiv{-0.4cm}
\label{fig:qual-comparison}
\end{figure*}

\arxiv{-0.3cm}
\paragraph{Photometric Error:} This energy encodes the fact that correspondences should have similar appearance across all images. In particular, for each pixel $\bm{p} \in P_i$ in the reference image, we compare its photometric value with that of the corresponding pixel in the target image:
\begin{align}
E_{\text{photo}, i}(\bm{\xi}; \cI) = & \sum_{\bp \in P_i} \alpha_\bp \rho( \cL^0 (\bp) - \cL^1(\bp^\prime)) %
\label{eq:dvo}
\end{align}
where  $\alpha_\bp \in \{0,1\}$ is an indicator function representing which pixel is an outlier. We refer the reader to section \ref{sec:inference} for a discussion on how to estimate $\alpha_p$.  $\bp$ is a pixel in the reference image and $\bp^\prime$ stands for the projected image coordinate on another image, given by inverse depth warping followed by a rigid transform $\bxi$. Specifically, 
\begin{equation}
\label{equ:warping}
\bp^\prime = \pi_\bK(\bxi \circ \pi_\bK^{-1}\left(\bp, \cD(\bp)\right))
\end{equation}
where $\pi_\bK(\cdot): \bbR^3 \rightarrow \bbR^2$ is the perspective projection function given  known intrinsic $\bK$ and $\pi_\bK^{-1}(\cdot, \cdot): \bbR^2\times \bbR \rightarrow \bbR^3$ is the inverse projection that convert a pixel and its associated disparity into a 3D point; $\bxi \circ \bx$  transforms a 3D point $\bx$ rigidly with transformation $\exp(\bxi) \bx$.  $\rho$ is a robust error function that improves the overall robustness by reducing the influence of outliers on the non-linear least squares problems. Following Sun \etal \cite{sun2010secrets}, we adopt the generalized Charbonnier function $\rho(x) = (x^2 + \epsilon^2)^\alpha$ as our robust function and set $\alpha = 0.45$ and $\epsilon = 10^{-5}$. Similar to \cite{sun2010secrets}, we observe the slightly non-convex penalty improves the performance in practice. %
\arxiv{-0.3cm}
\paragraph{Rigid Fitting:} This term encourages the estimated 3D rigid motion to be similar to the point-wise 3D motion obtained from the stereo and flow networks. Formally, given correspondences $\{(\bm{p}$, $\bm{q} = \bm{p} + F_\cL(\bp)) | \bm{p} \in P_i\}$ defined by the output of optical flow network and the disparity maps $\mathcal{D}^0, \mathcal{D}^1$, 
the energy measures rigid fitting error of $\bxi$: 
\[
E_{\text{rigid}, i}(\bm{\xi}; \cI) = \sum_{(\bp, \bq)} \alpha_\bp \rho( \bxi \circ \pi_\bK^{-1}\left(\bp, \cD^0(\bp)\right) - \pi_\bK^{-1}\left(\bq, \cD^1(\bq)\right)), \nonumber
\]
where $\bm{q} = \bm{p} + \mathcal{F}_\cL(\bm{p})$ and $\pi_\bK^{-1}$ denotes the inverse projection function, and $\rho$ is the same robust error function. %

\begin{figure*}[tb]
\arxiv{-0cm}
\centering
\setlength{\tabcolsep}{1pt}
\begin{tabular}{ccccc}
\includegraphics[width=0.19\linewidth]{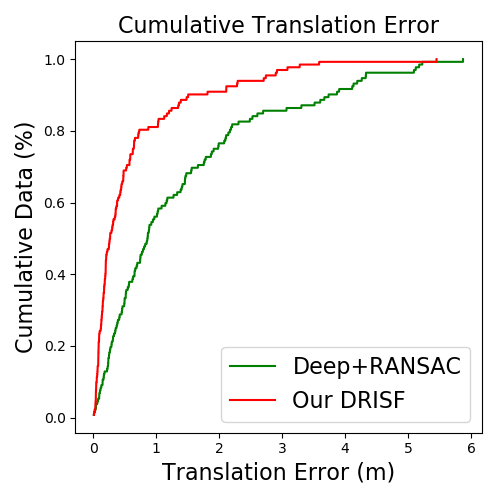}
&\includegraphics[width=0.19\linewidth]{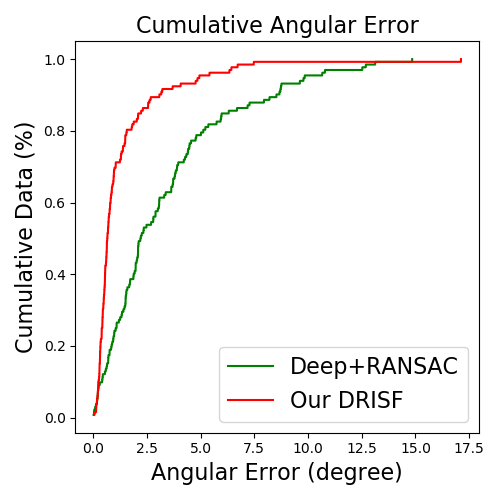}
&\includegraphics[width=0.19\linewidth]{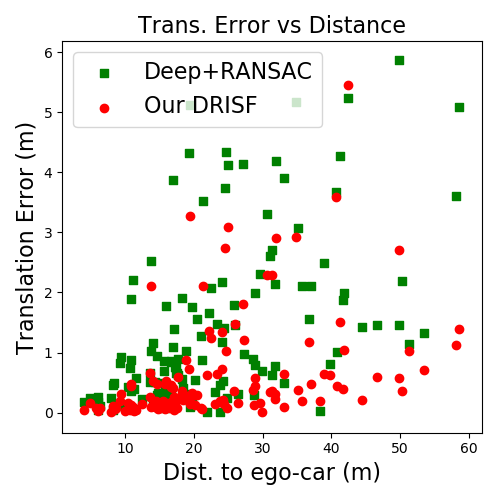}
&\includegraphics[width=0.19\linewidth]{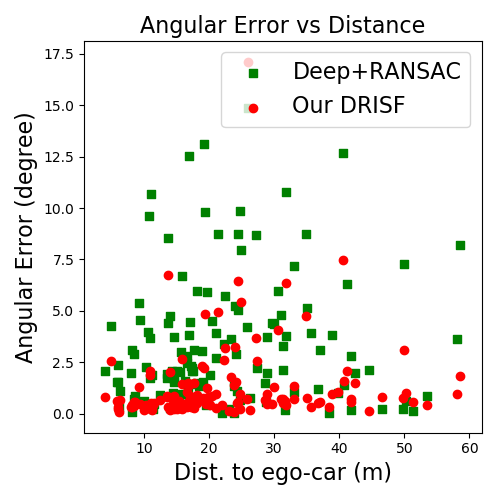}
&\includegraphics[width=0.19\linewidth]{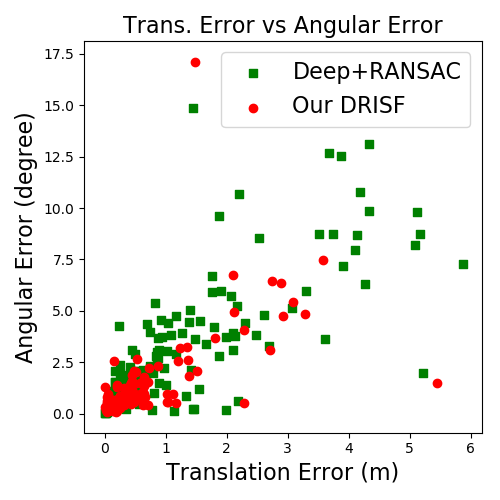}
\\\end{tabular}
\arxiv{-0.4cm}
\caption{\textbf{3D rigid motion analysis:} Over $80\%$ of the estimated 3D rigid motion has an error less than $1 m$ and $1.3^\circ$. Large errors often happen at farther distances where the vehicles are small and less points are observable.}
\arxiv{-0.6cm}
\label{fig:motion}
\end{figure*}

\begin{figure}[tb]
\centering
\setlength{\tabcolsep}{1pt}
\begin{tabular}{cc}
\includegraphics[width=0.48\linewidth]{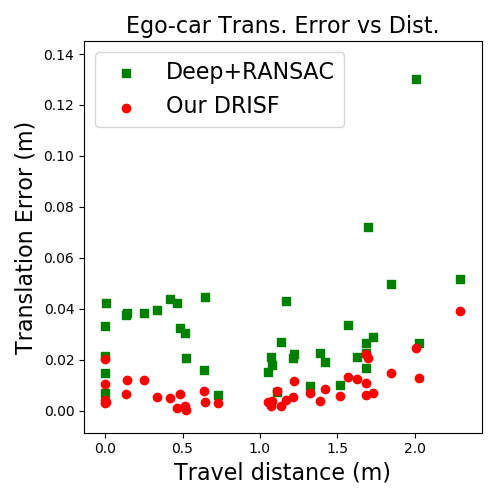}
&\includegraphics[width=0.48\linewidth]{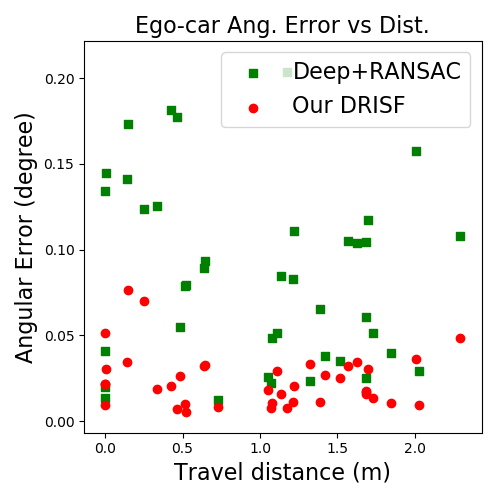}
\\\end{tabular}
\arxiv{-0.4cm}
\caption{\textbf{Odometry from background motion:} On average, our ego-car drifts $0.9 cm$ and $0.024^\circ$ every $1 m$ of drive.}
\arxiv{-0.4cm}
\label{fig:odometry}
\end{figure}

\arxiv{-0.4cm}
\paragraph{Flow Consistency:} This term encourages the projection of the 3D rigid motion to be close to the original flow estimation. 
This is achieved by measuring the difference between our optical flow net, and the structured rigid flow, which is computed by warping each pixel  using $\cD^0$ and the rigid motion $\bxi$.  %
\begin{align}
E_{\text{flow}, i}(\bm{\xi}; \cI) = \sum_{\bp \in P_i} \rho(\underbrace{(\bp^\prime - \bp)}_{\text{2D Rigid flow}} - \underbrace{\mathcal{F}_\cL(\bm{p})}_{\text{optical flow}})
\end{align}
where $\bp^\prime$ is the rigid warping function defined in Eq.~(\ref{equ:warping}), and $\rho$ is the same robust error function.   %

\arxiv{-0.2cm}
\subsection{Inference}
\label{sec:inference}
\arxiv{-0.2cm}
\paragraph{Uncertain Pixel Removal:} Due to viewpoint change, flow/stereo prediction errors, etc, the visual cues of some pixels are not reliable. For instance, pixels in one image may be occluded in another image due to viewpoint change. This motivates us to assign $\alpha_\bp$ to each pixel $\bp$ as an indication of outlier or not. Towards this goal, we first exclude pixels which are likely to be occluded in the next frame. Specifically, pixels are labeled as occluded if the warped 3D disparity of the second frame significantly differs from the disparity of the first frame. The intuition is that the disparity of a pixel cannot change drastically in real world due to the speed limit. We empirically set threshold to 30. Next, we employ the RANSAC scheme to fit a rigid motion for each instance. We only keep the inlier points and prune out the rest. Despite simple, we found this strategy very effective.
\arxiv{-0.4cm}
\paragraph{Initialization:} Due to the highly non-convex structure of the energy model, a good initialization is critical to achieve good performance. As previous step already prune out most unreliable points, we directly exploit the rigid motion obtained by RANSAC as our robust initial guess.%

\arxiv{-0.4cm}
\paragraph{Gaussian Newton Solver:} The energy function is non-convex but differentiable w.r.t. $\bxi$ defined over continuous space. In order to handle the robust function, we adopt an iterative reweighted least square algorithm \cite{candes2008enhancing}. For each iteration, we can rewrite the original energy minimization problem \addon{of each instance $i$} as a weighted sum of squares:
\[
\addon{\bxi^{(n+1)} = \arg\min_{\bxi} E_{\text{total}, i}(\bxi) = \arg\min_{\bxi} \sum_{\text{Eng}} w_i(\bxi^{(n)}) r_i^2 (\bxi^{(n)}),}
\]
\addon{where $r$ denotes the residual function, $w$ reweights each sample based on the robust function $\rho$, and Eng refers to summing over the energy terms.}
We employ Gaussian-Newton algorithm to minimize the function. Thus we have 
\begin{align}
\label{equ:solver}
\bxi^{(n+1)} = -(\bJ^T \bW \bJ)^{(-1)} \bJ^T \bW r(\bxi^{(n)})  \circ \bxi^{(n)}
\end{align}
where $\circ$ is a pose composition operator and $\bJ = \frac{\delta r (\bm{\epsilon} \circ \bxi^{(n)})}{\delta \bm{\epsilon}}|_{\bm{\epsilon} = 0}$.
In practice, we unroll the inference steps as a recurrent neural network and define its computation graph as in Eq.~(\ref{equ:solver}). The full pipeline including the matrix inverse is differentiable. Please refer to the supp. material for the derivation of the Jacobian matrix of each term \addon{and more details on the Gaussian-Newton solver}.

\paragraph{Final Scene Flow Prediction:}
Given the final rigid motion estimation for each instance $\bxi^\ast_i$, we are able to compute the dense instance-wise rigid scene flow. Our scene flow consists of three component, namely the first frame's stereo $\cD^0$, warped stereo to second frame $\cD^{\mathrm{warp}}$ as well as the instance-wise rigid flow estimation $\cF^{\mathrm{rigid}}$. 
Specifically, for each point $\bp$ we have:
\begin{align} 
\cD^0(\bp) &= \cD^0(\bp) \\ \nonumber
 \cD^{\mathrm{warp}}(\bp) &= z_\bK(\bxi^\ast_{\cS_\cL^0(\bp)} \circ \pi_\bK^{-1} (\bp, \cD^0(\bp)))\\ \nonumber
 \cF^{\mathrm{rigid}}(\bp) &= \bp^\prime - \bp =  \pi_\bK(\bxi \circ \pi_\bK^{-1}\left(\bp, \cD^0(\bp)\right)) - \bp
\end{align}
where  $z_\bK(\cdot)$ computes the disparity of the 3D point; $\pi_\bK^{-1}$ is the inverse projection function; and $\bxi \circ \bx$ transforms a 3D point $\bx$ using the rigid motion $\bxi$.

\subsection{Learning}
The whole deep structured network can be trained  end-to-end. In practice, we train our instance segmentation, flow estimation, and stereo estimation module respectively through back-propagation. To be more specific, Mask R-CNN model is pre-trained on Cityscapes and fine-tuned on KITTI. The loss function includes ROI classification loss, box regression loss as well as the mask segmentation loss. PSM-Net is pre-trained on Scene Flow \cite{mayer2016large} and fine-tuned on KITTI with L1 regression loss. PWC-Net is pre-trained on FlyingChairs \cite{fischer2015flownet} and FlyingThings \cite{mayer2016large} then fine-tuned over KITTI, with weighted L1 regression loss. 
\section{Experiments}

In this section we first describe the  experimental setup. Next we evaluate our model based on pixel-level scene flow metric and instance-level rigid motion metric. Finally we comprehensively study the characteristic of our model. %

\begin{table*}[tb]
\arxiv{-0.7cm}
\centering
\begin{tabular}{ccccccc}
\specialrule{.2em}{.1em}{.1em}
\multicolumn{3}{c}{Employed energy}&\multicolumn{4}{c}{Background outliers (\%)}\\
${E}_{pho}$& ${E}_{flow}$&${E}_{rigid}$ &D1&{D2} &{Fl}  &{SF}\\
\hline
\checkmark& & &1.92 &2.69 &{\bf3.71} &{\bf4.30}  \\

& \checkmark& \checkmark &1.92 & {\bf2.56}& 4.72& 5.28\\
\checkmark& \checkmark & \checkmark &1.92 &{\bf2.56} &4.63 &5.21\\
\specialrule{.1em}{.05em}{.05em}
\end{tabular}
\begin{tabular}{ccccccc}
\specialrule{.2em}{.1em}{.1em}
\multicolumn{3}{c}{Employed energy}&\multicolumn{4}{c}{Foreground outliers (\%)}\\
${E}_{pho}$& ${E}_{flow}$&${E}_{rigid}$ &D1&{D2} &{Fl}  &{SF}\\
\hline
\checkmark& & &1.70 &{\bf4.25} &7.57 &9.00 \\

& \checkmark& \checkmark &1.70 & 4.58 & 6.98 & 8.67\\
\checkmark& \checkmark & \checkmark &1.70 &4.56 &{\bf6.73} &{\bf8.39}\\
\specialrule{.1em}{.05em}{.05em}
\end{tabular}
\arxiv{-0.2cm}
\caption{\textbf{Contributions of each energy:} As foreground objects sometimes are texture-less and have large displacement, simple photometric term is not enough. In contrast, background is full of disriminative cues. Simple photometric error would suffice. Adding extra terms will introduce noises and hurt the performance. Please refer to the supp. material for full table.}
\arxiv{-0.5cm}
\label{tab:ablation-energy}
\end{table*}

\subsection{Dataset and Implementation Details}

\paragraph{Data:} We validate  our approach on the KITTI scene flow dataset \cite{menze2015object}. The dataset consists of 200 sets of  training images and 200 sets of test images, captured on real world driving scenarios. 
Following \cite{chang2018pyramid}, we divide the training data into \emph{train}, \emph{val} splits based on the 4:1 ratio.

\paragraph{Implementation details:} For foreground objects, we use all energy terms. The weights are set to 1. For background, we only use photometric term (see ablation study). We run RANSAC 5 times and use the one with lowest mean energy as initialization. 
We unroll the GN solver for 50 steps. The solver terminates early if the energy reaches plateau. In practice, best energy are often reached within 10 iterations.  %

\begin{table}[tb]
\centering
\begin{tabular}{lcccc}
\specialrule{.2em}{.1em}{.1em}
Methods &{D1-all} &{D2-all} &{Fl-all} &{SF-all}\\
\hline
PSM + PWC &1.89& (47.0)&11.0 & (50.8)\\
Deep+RANSAC &1.89 & {\bf2.75}  & 7.65 & 8.26 \\ 
Our Full \Abbrev~&1.89  & 2.89  & {\bf4.10} & {\bf4.84}\\
\specialrule{.1em}{.05em}{.05em}
\end{tabular}
\arxiv{-0.2cm}
\caption{\textbf{Improvement over original flow/stereo estimation on validation set:} The numbers in parenthis are obtained by simply warping the disparity output with optical flow, without interpolation, occlusion handling, etc.}
\arxiv{-0.3cm}
\label{tab:before-after}
\end{table}

\begin{table}[tb]
\centering
\scalebox{0.77}{
\begin{tabular}{lccc}
\specialrule{.2em}{.1em}{.1em}
Module & Stereo & Optical Flow & Segmentation\\
\hline
Inference time & \cellcolor{blue!25}409 ms / pair & \cellcolor{blue!25}30 ms / pair & \cellcolor{blue!25}251 ms / pair  \\ 
\hline
Module & RANSAC & GN Solver & Total\\
\hline
Inference time &\cellcolor{orange!25} 93 ms / instance &\cellcolor{green!25} 244 ms / instance & 746 ms / pair\\
\specialrule{.1em}{.05em}{.05em}
\end{tabular}
}
\arxiv{-0.2cm}
\caption{\textbf{Runtime analysis.} Modules within each building block can be executed in parallel (see text for more details).}%
\arxiv{-0.6cm}
\label{tab:runtime}
\end{table}

\subsection{Scene Flow Estimation}

\paragraph{Comparison to the state-of-the-art:} We compare our approach against the leading methods on the benchmark\footnote{As the validation performance of our PWC-Net (fine-tuned on 160 images) performs slightly worse than the official one (fine-tuned on all 200 images), we use their weights instead when submitting to the benchmark. All other settings remain intact. We thank Deqing Sun for his help.}: ISF \cite{behl2017bounding}, PRSM \cite{vogel20153d}, OSF+TC \cite{artnerobject}, SSF \cite{ren2017cascaded}, OSF \cite{menze2015object}, and CSF \cite{lv2016continuous}. Note that in addition to the standard two adjacent frames, PRSM and OSF+TC  rely on extra temporal frames.
As shown in Tab. \ref{tab:quant}, our approach (\Abbrev) outperforms all previous methods by a significant margin in both runtime and outliers ratio. It achieves state-of-the-art performance on almost every entry.  \Abbrev~reduces the D1 outliers ratio by \textbf{43\%}, the D2 outliers ratio by \textbf{32\%}, and the flow outliers ratio by \textbf{24\%}. Comparing to  ISF model \cite{behl2017bounding}, our scene flow error is \textbf{22\%} lower and our runtime is \textbf{800} times faster. Fig. \ref{fig:perf-vs-runtime} compares the performance and runtime of all methods.

\paragraph{Qualitative results:} To better understand the pros and cons of our approach, we visualize a few scene flow results on test set in Fig. \ref{fig:qual-comparison}.
Scene flow estimation is challenging in these scenarios due to  large vehicle motions, texture-less regions, occlusion, and illumination variation.
For the leftmost image, prior methods fail to estimate the vehicle's motion and adjacent area due to the sun reflection and occlusion. The saturated, high intensity pixels hinder photometric based approaches \cite{menze2015object} from matching accurately. %
With the help of detection and segmentation, ISF \cite{behl2017bounding} is able to improve the foreground estimation. Yet it still fails at the occluded background. In comparison, our approach is robust to illumination changes and is able to handle the occlusion by effectively separating the vehicle from the background. It can also accurately estimate the motion of the small car far away, as well as those of the traffic sticks aside. 
As we only train our Mask R-CNN on vehicles, it fails to segment the train and hence the failure of our model.
For the middle image, the texture-less car has a large displacement and is occluded in the second frame. While previous approaches failed substantially, our method is able to produce accurate motion estimation through the inferred flow and disparity of the remaining non-occluded part. The middle failure mode is again due to the inaccurate segmentation.

\begin{figure*}[tb]
\arxiv{-0.7cm}
\centering
\setlength{\tabcolsep}{1pt}
\arxivscale{0.95}{
\begin{tabular}{cc|cc}
Before (PWC) & After (\Abbrev) & Before (PSM+Warp) & After (\Abbrev)\\
\includegraphics[width=0.24\linewidth]{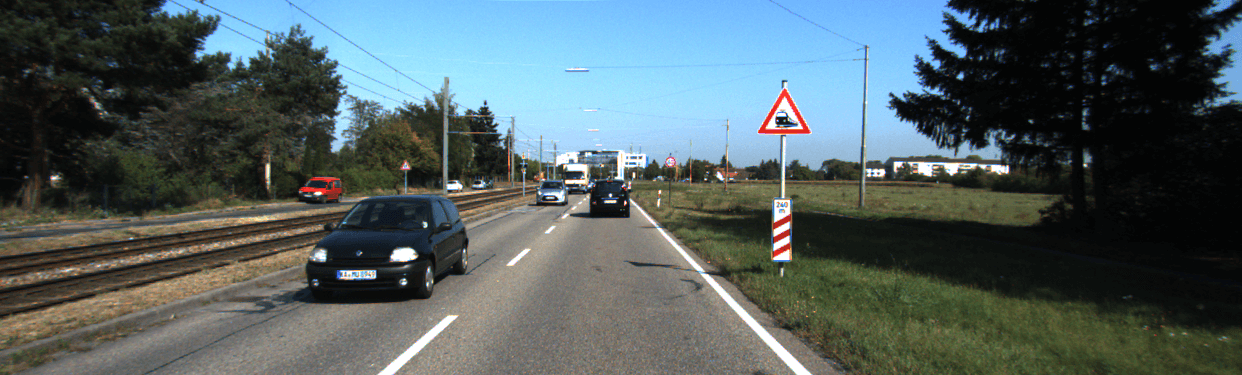}
&\includegraphics[width=0.24\linewidth]{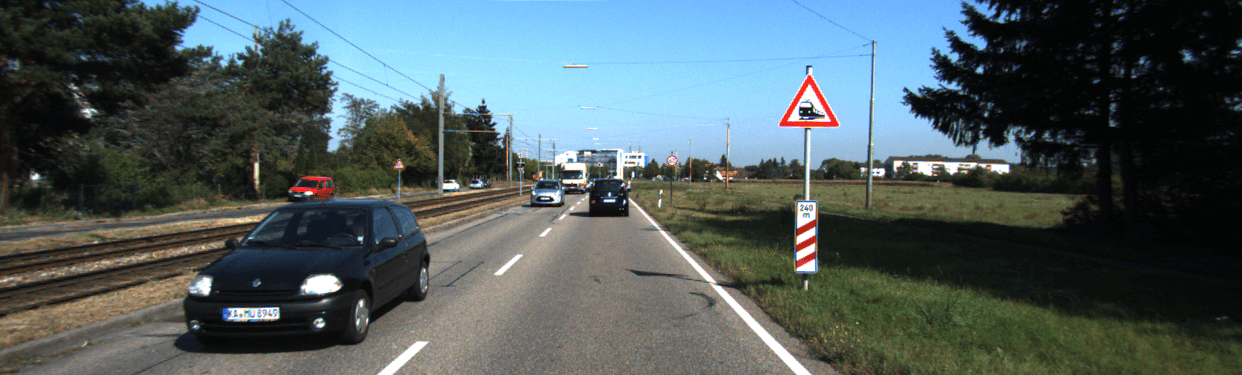}
&\includegraphics[width=0.24\linewidth]{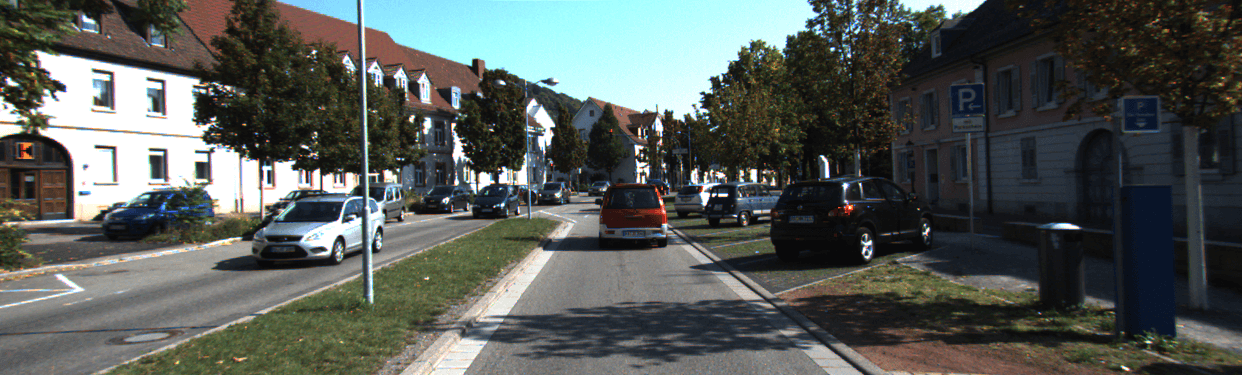}
&\includegraphics[width=0.24\linewidth]{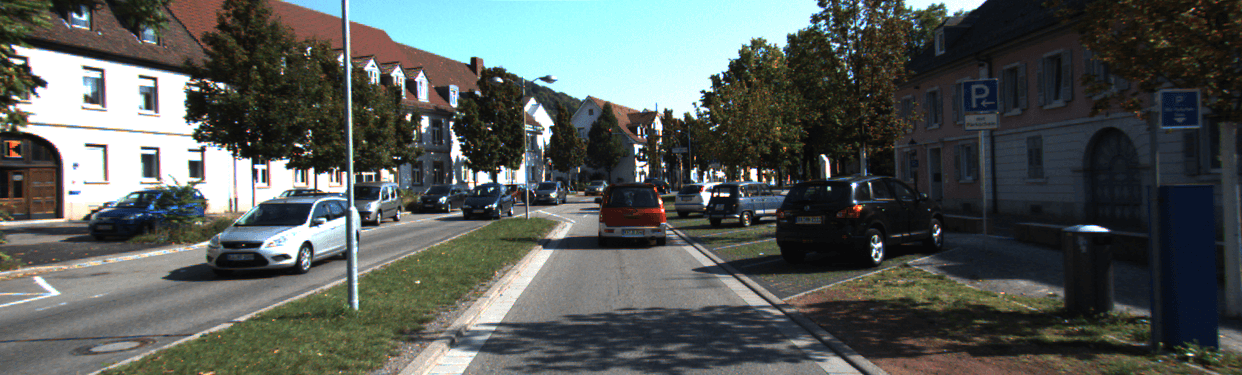}\\
\includegraphics[width=0.24\linewidth]{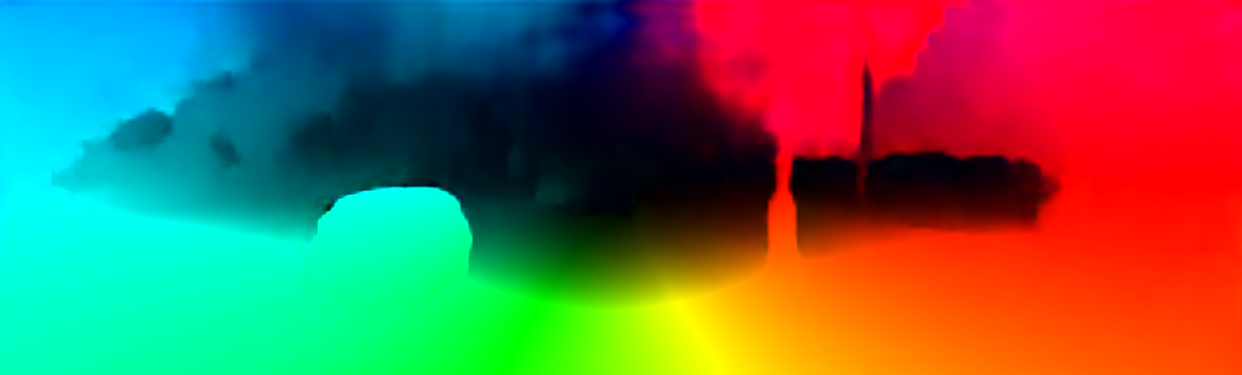}
&\includegraphics[width=0.24\linewidth]{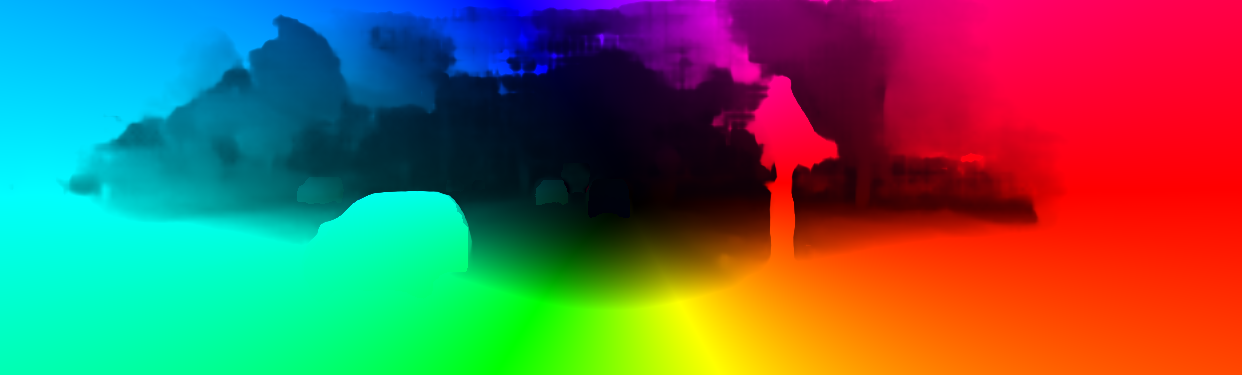}
&\includegraphics[width=0.24\linewidth]{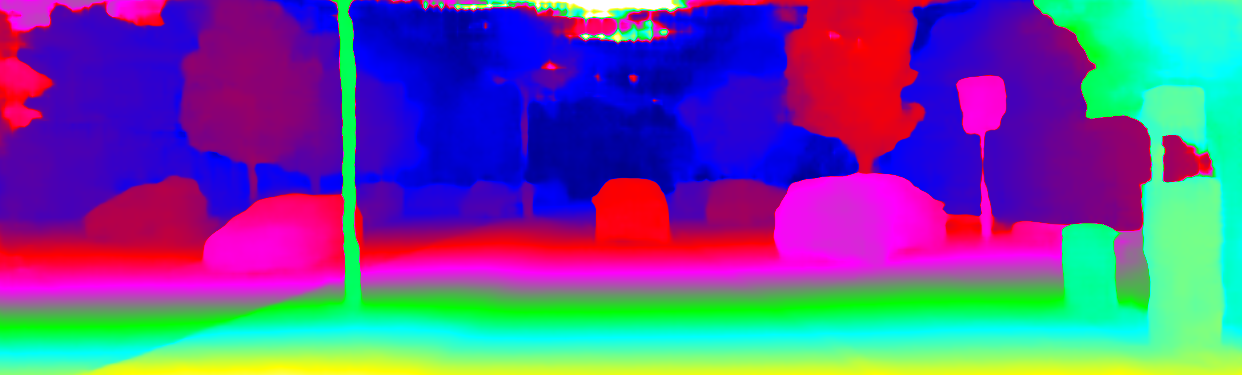}
&\includegraphics[width=0.24\linewidth]{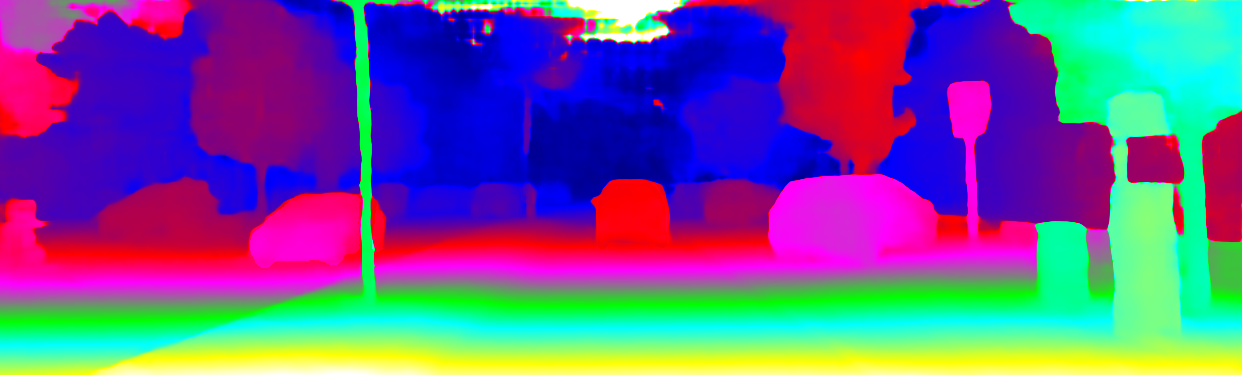}\\
\includegraphics[width=0.24\linewidth]{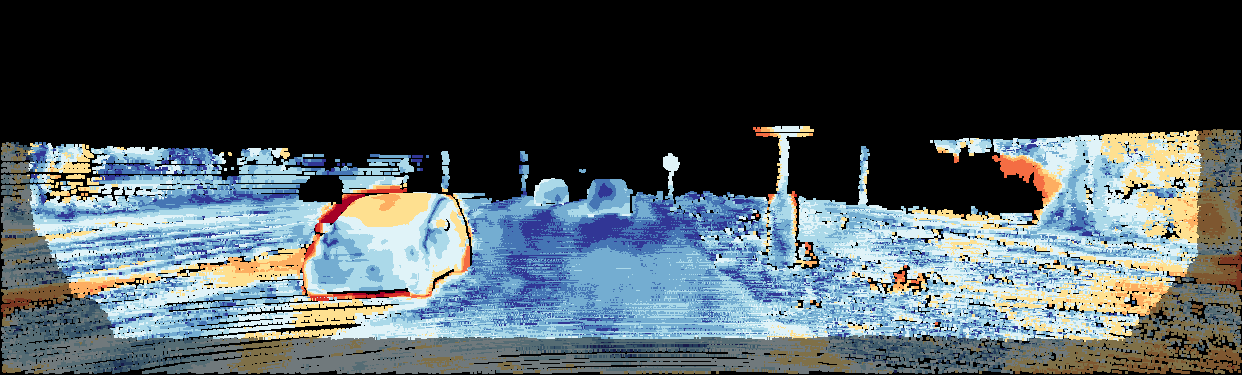}
&\includegraphics[width=0.24\linewidth]{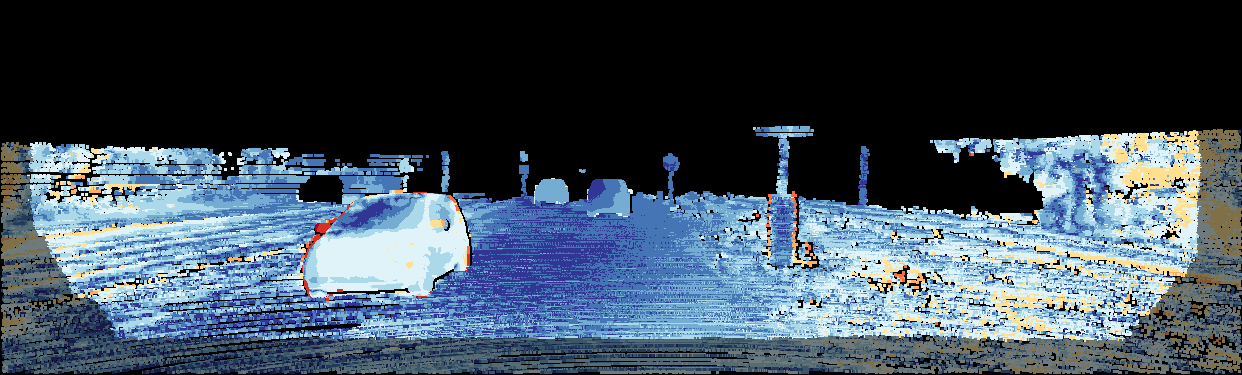}
&\includegraphics[width=0.24\linewidth]{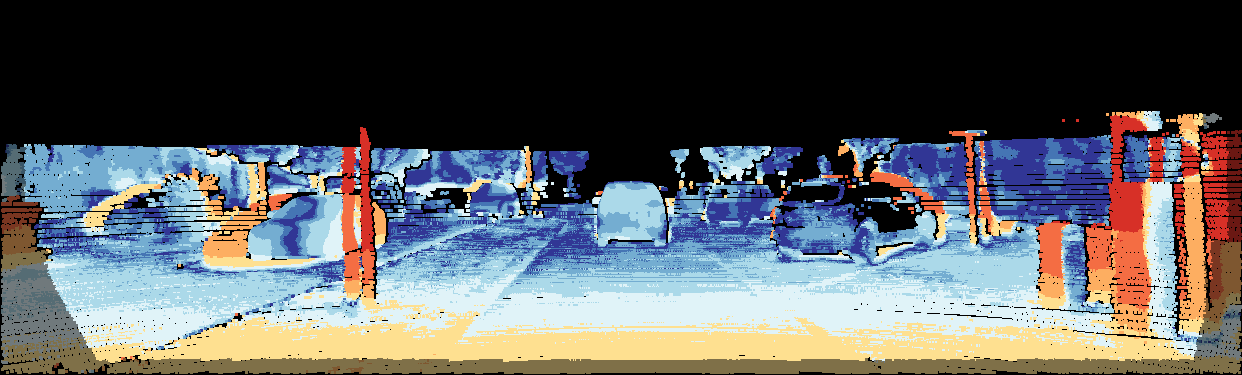}
&\includegraphics[width=0.24\linewidth]{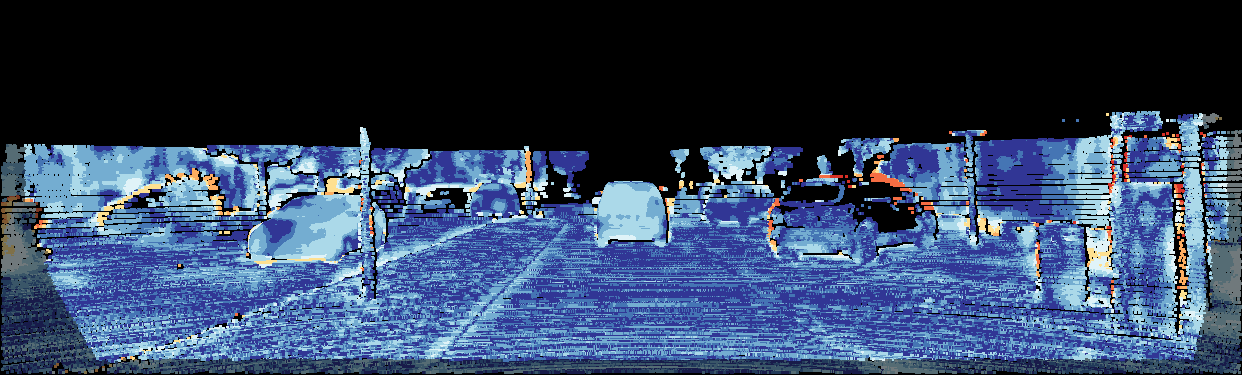}\\
\multicolumn{4}{c}{\includegraphics[width=0.97\linewidth]{img/kitti_test/error_bar.png}}
\\\end{tabular}
}
\arxiv{-0.3cm}
\caption{\textbf{Improvement over original flow/stereo:} \Abbrev~improves the overall performance. It is especially effective on texture-less regions (\eg window of the black car on the left) and occluded areas (right).}
\arxiv{-0.5cm}
\label{fig:before-after}
\end{figure*}

\subsection{3D Rigid Motion Estimation}
We now evaluate how good our \Abbrev~model is at estimating the 3D rigid motion. 
Towards this goal, we exploit the ground truth optical flow, disparity, and instance segmentation provided in the KITTI scene flow dataset to fit a least square rigid motion for each object instance in order to create the ground truth rigid motion. 

\paragraph{Curating KITTI scene flow: } During fitting, we discover two critical issues with KITTI: first, there are mis-alignments between GT flow/disparity and GT segmentation. Second, the scale fitting of the same 3D CAD model employed to compute ground truth changes   sometimes  across frames.
The first issue is due to the fact that the GT are collected via different means and thus not consistent. While the GT flow and GT disparity are obtained from the fitted 3D CAD models, the GT segmentation are based on human annotation. 
To address this, we first use the GT segmentation mask to define each object instance. We then fit a rigid motion using the GT flow and GT disparity of each instance via least squares. Since some boundary pixels may be mis-labeled by the annotators, for each pixel around the boundary we search if there are other instances in the surrounding area, and if there are, we transform the pixel with their rigid motion. If their rigid motion better explains the pixel's 3D movement, \emph{i.e.} the 3D distance is closer, then we assign the pixel to that instance. At the end, we perform the least square fitting again with the new pixel assignment.
Unfortunately, even after re-labeling, there are still a few vehicle instances where the rigid motion cannot be explained. After careful diagnose, we notice that this is because the scale of the CAD model changes across frames. To verify our hypothesis, we compute the eigen decomposition for the same instance across frames. Ideally if the scale of the instance does not change much, the eigen value should be roughly the same. Yet we discover a few examples where the largest eigen value changes by $7\%$. 
We simply prune those instances as the GT is  not accurate. 

\paragraph{3D Motion evaluation:} \cbd{Most scene flow methods are pixel-based or adopted a piece-wise rigid setting. It is unclear how to aggregate their estimation into instance-based motion model without affecting their performance. In light of this,} we exploit the motion initialization of our GN Solver as baseline. 
We take the output of the deep nets and apply RANSAC to find the best rigid motion. We denote it as Deep+RANSAC. As shown in Tab. \ref{tab:before-after}, this baseline is very competitive. Its performance is comparable to, or even better than prior state-of-the-art. %
We evaluate our motion model based on translation error and angular error. As shown in Fig. \ref{fig:motion}, over $80\%$ of the vehicles have translation error less than $1 m$ and angular error less than $1.3^\circ$. Furthermore, most vehicles with translation error larger than $1 m$ is at least $20 m$ away. 
In general, both error slightly increase with  distance. This is expected as the farther the vehicle is, the less observations we have. The translation error and angular error are also strongly correlated.

\paragraph{Visual odometry:} 
The odometry of the `ego-car' can be computed by estimating the background movement. As a proof-of-concept, we compute the per frame odometry error on the validation images. On average our motion model drifts $0.09 m$ and $0.24^\circ$ every $10 m$. %
Fig. \ref{fig:odometry} shows the detailed odometry error w.r.t. the travel distance. We note that the current result is without any pose filter, loop closure, etc. We plan to exploit this direction further in the future. %

\subsection{Analysis}
\label{sec:analysis}
\paragraph{Ablation study:} To understand the effectiveness of each energy term on background and foreground objects, we evaluate our model with different energy combinations. As shown in Tab. \ref{tab:ablation-energy}, best performance is achieved for foreground objects when using all energy terms, while for background the error is lowest when employing only photometric term. This can be explained by the fact that vehicles are often texture-less, and sometimes have large displacements. If we only employ photometric term, it will be very difficult to establish correspondences and handle drastic appearances changes. With the help of flow  and rigid term, we can guide the motion and reduce such effect, and deal with occlusions. In contrast, background is full of discriminative textures and has relatively small motion, which is ideal for photometric term. Adding other terms may introduce extra noise and degrade the performance. %

\arxiv{-0.3cm}
\paragraph{Comparison against original flow/disparity:} 
Through exploiting the structure between visual cues and occlusion handling, our model is able to improve the performance both quantitatively (Tab. \ref{tab:before-after}) and qualitatively (Fig. \ref{fig:before-after}). \cbd{The object motion estimation is better, the boundaries are sharper, and the occlusion error is greatly reduced, suggesting that incorporating prior knowledge, such as pixels of same instance should have same rigid motion, into the model is crucial for the task.}
\arxiv{-0.3cm}

\paragraph{Potential improvement} To understand the potential gain we may enjoy when improving each module, we sequentially replace the input to our solver with ground truth, one by one, and evaluate our model. %
Replacing D1 and flow with GT reduce the scene flow error rate by $8\%$ and $21\%$ respectively, while substituting GT for segmentation does not improve the results. %
This suggests that there are still space for flow and stereo modules to improve.

\paragraph{Runtime analysis} 
We benchmark the runtime of each component in the model during inference  in Tab. \ref{tab:runtime}. The whole inference pipeline can be decomposed into three sequential stages: visual cues extraction, occlusion reasoning, and optimization. As modules within the same stage are independent, they can be executed in parallel. \addon{Furthermore, modern self-driving vehicles are equipped with multiple GPUs.} The runtime for each stage is thus the max over all parallel modules. \addon{In practice, we exploit two Nvidia 1080Ti GPUs to extract the visual cues: one for PSM-Net, and the other for Mask R-CNN and PWC-Net.} \cbd{Currently, the stereo module takes more than $50\%$ of the overall time. This is largely due to the 3D CNN cost aggregation and the stacked hourglass refinement. In the future, we plan to investigate other faster yet reliable stereo networks. The runtime of the GN solver depends highly on the number of steps we unroll and the number of points we consider. Please refer to the supp. material for detailed analysis.}

\paragraph{Limitations:} \Abbrev~has two main limitations: first, it heavily depends on the performance of the segmentation network. If the segmentation module fails to detect a vehicle, the vehicle will be treated as background and assigned an inverse ego-car motion. In this case, the 3D motion might be completely wrong, even if the optical flow network accurately predicts its flow. In the future we plan to address this by jointly reasoning about instance segmentation and scene flow. 
Second, the current energy functions are highly flow centric. Only the photometric term is independent of flow. If the optical flow network completely failed, it would be difficult for the solver to recover the correct motion. One possible solution is thus adding more flow-invariant energy terms, such as instance association between adjacent frames. 
\section{Conclusion}
\vspace{-0.2cm}
In this paper we develop a novel deep structured model for 3D scene flow estimation. We focus on the self-driving scenario where the motion of the scene can be composed by estimating the 3D rigid motion of each actor. We first exploit deep learning to extract visual cues for each instance. Then we employ multiple geometry based energy functions to encode the structural geometric relationship between them. %
Through optimizing the energy function, we can reason the 3D motion of each traffic participant, and thus scene flow. All operations, including the Gassian-Newton solver, are done in GPU.
Our method acheives state-of-the-art performance on the KITTI scene flow dataset. It outperforms all previous methods by a huge margin in both runtime and accuracy. Comparing to prior art, \Abbrev~is $22\%$ better while being two to three orders of magnitude faster.

{
\footnotesize
\bibliographystyle{ieee}
\bibliography{egbib}
}

\end{document}